\documentclass[finalnew]{agujournal2019}

\usepackage{url} %this package should fix any errors with URLs in refs.
\usepackage{lineno}
\usepackage[inline]{trackchanges} %for better track changes. finalnew option will compile document with changes incorporated.
\usepackage{soul}
\usepackage{booktabs}

 \usepackage{graphicx}
\usepackage{amsmath,amssymb,amsfonts}%
 \usepackage{amsthm}%
\addeditor{Alex}

\draftfalse

\journalname{Earth and Space Science}

\begin{document}

%% ------------------------------------------------------------------------ %%
%  Title
%
% (A title should be specific, informative, and brief. Use
% abbreviations only if they are defined in the abstract. Titles that
% start with general keywords then specific terms are optimized in
% searches)
%
%% ------------------------------------------------------------------------ %%

% Example: \title{This is a test title}

\title{Deep Autoencoders for Unsupervised Anomaly Detection
in Wildfire Prediction}

%% ------------------------------------------------------------------------ %%
%
%  AUTHORS AND AFFILIATIONS
%
%% ------------------------------------------------------------------------ %%

% Authors are individuals who have significantly contributed to the
% research and preparation of the article. Group authors are allowed, if
% each author in the group is separately identified in an appendix.)

% List authors by first name or initial followed by last name and
% separated by commas. Use \affil{} to number affiliations, and
% \thanks{} for author notes.
% Additional author notes should be indicated with \thanks{} (for
% example, for current addresses).

% Example: \authors{A. B. Author\affil{1}\thanks{Current address, Antartica}, B. C. Author\affil{2,3}, and D. E.
% Author\affil{3,4}\thanks{Also funded by Monsanto.}}

\authors{İrem Üstek\affil{1}, Miguel Arana-Catania\affil{1}, Alexander Farr\affil{2}, Ivan Petrunin\affil{1}}

\affiliation{1}{School of Aerospace, Transport and Manufacturing, Cranfield University, UK}
\affiliation{2}{Spire Global, Inc., Boulder, CO, United States}
% \affiliation{3}{Third Affiliation}
% \affiliation{4}{Fourth Affiliation}

%\affiliation{=number=}{=Affiliation Address=}
%(repeat as many times as is necessary)

%% Corresponding Author:
% Corresponding author mailing address and e-mail address:

% (include name and email addresses of the corresponding author.  More
% than one corresponding author is allowed in this LaTeX file and for
% publication; but only one corresponding author is allowed in our
% editorial system.)

% Example: \correspondingauthor{First and Last Name}{email@address.edu}

\correspondingauthor{Miguel Arana-Catania}{miguel.aranacatania@cranfield.ac.uk}

%% Keypoints, final entry on title page.

%  List up to three key points (at least one is required)
%  Key Points summarize the main points and conclusions of the article
%  Each must be 140 characters or fewer with no special characters or punctuation and must be complete sentences

% Example:
% \begin{keypoints}
% \item	List up to three key points (at least one is required)
% \item	Key Points summarize the main points and conclusions of the article
% \item	Each must be 140 characters or fewer with no special characters or punctuation and must be complete sentences
% \end{keypoints}

\begin{keypoints}
\item  This research addresses a gap in the literature by exploring the advantages of unsupervised learning in wildfire prediction
\item This study focuses on Australia using a dataset comprising historical weather and normalized difference vegetation index
\item Our findings demonstrate the effectiveness of unsupervised learning using deep autoencoders for anomaly detection with an F1-score of 0.74
\end{keypoints}

%% ------------------------------------------------------------------------ %%
%
%  ABSTRACT and PLAIN LANGUAGE SUMMARY
%
% A good Abstract will begin with a short description of the problem
% being addressed, briefly describe the new data or analyses, then
% briefly states the main conclusion(s) and how they are supported and
% uncertainties.

% The Plain Language Summary should be written for a broad audience,
% including journalists and the science-interested public, that will not have 
% a background in your field.
%
% A Plain Language Summary is required in GRL, JGR: Planets, JGR: Biogeosciences,
% JGR: Oceans, G-Cubed, Reviews of Geophysics, and JAMES.
% see http://sharingscience.agu.org/creating-plain-language-summary/)
%
%% ------------------------------------------------------------------------ %%

%% \begin{abstract} starts the second page

\begin{abstract}
 Wildfires pose a significantly increasing hazard to global ecosystems due to the climate crisis. Due to its complex nature, there is an urgent need for innovative approaches to wildfire prediction, such as machine learning. This research took a unique approach, differentiating from classical supervised learning, and addressed the gap in unsupervised wildfire prediction using autoencoders and clustering techniques for anomaly detection. Historical weather and normalised difference vegetation index datasets of Australia for 2005 – 2021 were utilised. Two main unsupervised approaches were analysed. The first used a deep autoencoder to obtain latent features, which were then fed into clustering models, isolation forest, local outlier factor and one-class SVM for anomaly detection. The second approach used a deep autoencoder to reconstruct the input data and use reconstruction errors to identify anomalies. Long Short-Term Memory (LSTM) autoencoders and fully connected (FC) autoencoders were employed in this part, both in an unsupervised way learning only from nominal data. The FC autoencoder outperformed its counterparts, achieving an accuracy of 0.71, an F1-score of 0.74, and an MCC of 0.42. These findings highlight the practicality of this method, as it effectively predicts wildfires in the absence of ground truth, utilising an unsupervised learning technique.
\end{abstract}

\section{Introduction}\label{intro}
Forests are essential to maintaining the planet's ecological equilibrium, but they face threats from fires caused by
both natural phenomena and human activities \cite{qiang}. These wildfires are catastrophic
events that negatively impact the environment, the economy, and the natural resources \cite{meddour2015wildfires,sayad2019predictive}.  On the human front, they can lead to loss of life,
respiratory ailments, and other health challenges. At the same time, economically, they can devastate communities,
affecting livelihoods and leading to billions in damages. The impact on wildlife is also severe, threatening a diverse
range of species with effects such as respiratory distress, altered behaviour, and even the threat of extinction. Simultaneously, the natural
environment suffers degradation through soil erosion, loss of biodiversity, and the destruction of essential ecosystems such as forests \cite{sanderfoot2022review}. 

In recent years,
the frequency of wildfires has surged, becoming a global concern that has drawn attention from various fields of study
\cite{abid2021survey}.  A very recent example would be the wildfire episodes encompassing 25 million acres in Canada in 2023 \cite{Canada}. The growing threat of wildfire disasters is primarily attributed to climate change. These changes, characterised by higher temperatures and extended dry periods, have led to unprecedented bushfire activities, particularly within Australia \cite{vardoulakis2020lessons}.

In 2019, Australia, the focus area of this
research, encountered its hottest year on record, with the annual national mean temperature surpassing the
average by 1.52°C. It also experienced its driest year on record, with notable heatwaves occurring in January and
December \cite{yu2020bushfires}. Between 2019 and 2020, Australia witnessed one of the largest wildfires in modern record. A minimum area of 46 million acres of land was burnt
\cite{sulova2020exploratory}. The 2019-20 wildfire is noted for having the strongest measures on record in terms of both magnitude and intensity \cite{zhang2021analysis}. Regarding their impacts on climate change and the environment, these
fires directly claimed 33 human lives and over one billion indigenous animals \cite{norman2021apocalypse}. Furthermore, 417
individuals succumbed to smoke inhalation, and it is believed that 80\% of Australia's population was directly or
indirectly affected by the fires \cite{borchers2020unprecedented,norman2021apocalypse,ogie2022twitter}. Moreover, projections indicate that such wildfires will continue to occur due to multiple factors. Based on scientific estimates by the
United Nations, extreme wildfires could increase by as much as 50\% by the century's end, spurred by the climate crisis
and poor land management \cite{qiang}. 

Simulating wildfires is challenging due to the complex interactions of weather conditions, fuel setups, and
unpredictable fire movements. The wide geographical and time scales involved make this difficulty even greater, making
it a problem that requires significant computational resources \cite{sayad2019predictive}. The difficulty in predicting
wildfires emphasises the urgent need for innovative approaches, such as machine learning (ML) techniques. Indeed,
numerous studies have been conducted to predict or detect wildfires in various scenarios using supervised learning
techniques, such as support vector machines (SVM), random forests (RF), artificial neural networks (ANN), decision
trees (DT), and more \cite{abid2021survey,pang2022forest,perez2021machine,shmuel2022global,sulova2020exploratory}. However, unsupervised learning in wildfire detection has not received much attention from researchers in the prediction phase, and only has been utilized in exploratory data analysis \cite{sayad2019predictive}. This research aims to bridge this gap by leveraging the advantages of unsupervised learning, such as accessibility to more data without the need for specific wildfire datasets, and the capability for broader and more flexible analysis without requiring labeled data. Advancements in unsupervised learning, exemplified by generative adversarial networks (GANs), a type of unsupervised autoencoder, have been shown to have potential for rapid, real-time applications in contexts such as wildfire detection \cite{zenati2018adversarially}. This underscores the broader applicability and benefits of unsupervised learning techniques in this field. The benefits of using unsupervised learning methods are also shown in multiple and varied domains, requiring less computational power, less time consumption and use of less amount of data while yielding relatively good quality results in near-real time \cite{nhangumbe2023supervised}, reducing the data processing effort by up to 97\% while being reliable enough for the task requirements \cite{lehr2021supervised}, showing strong scalability and interpretability  \cite{chang2020landslide}, achieving the task requirements while reducing the training cost \cite{tuo2023supervised}, and sometimes producing results on par with supervised techniques \cite{haar2019comparison, pradhan2021computational}.

The study will specifically narrow its focus to Australia. This area is particularly prone to wildfires and has a
significant history in this domain, making it a valuable subject for examination. The research aims to predict future wildfires based on anomaly detection before they happen by employing unsupervised
learning techniques, using deep autoencoders and clustering. The dataset employed for this project is the combination of two separate datasets: historical weather and Normalised Difference Vegetation Index (NDVI) \cite{IBM}. It employs unsupervised learning techniques, specifically deep autoencoders and clustering, to predict wildfires through anomaly detection, utilizing a unique dataset comprising historical weather and normalized difference vegetation index data. The techniques employed are some of the most common unsupervised methods for anomaly detection \cite{li2021deep,cheng2021improved}. This research addresses a gap in the literature by exploring the potential advantages of unsupervised learning in wildfire prediction, such as its reliance on readily available data and reduced computational demands \cite{nhangumbe2023supervised, lehr2021supervised, tuo2023supervised}. In addition to primary contributions, this study also addresses challenges in wildfire prediction and recommends ways to enhance predictive models in this domain. 

The remaining sections of this paper are organised as follows: Section \ref{related_work} offers a literature review; Section \ref{method} provides details about the study
area, dataset, and methodology; Section \ref{results} presents results and findings; Section \ref{discussion} discusses the results; and Section
\ref{concl} provides conclusions.

\section{Related Work}\label{related_work}

In this section we review relevant research to our project, beginning with a general study on wildfire prediction using machine learning, followed by specific reviews of the techniques employed in our work: anomaly detection, autoencoders, and the combination of both.

\subsection{Wildfire Prediction Using ML Models}
Various ML techniques and features have been applied and researched regarding forest fire prediction and detection. The most common and popular ML models used are the artificial neural networks (ANN), support vector machines (SVM), random forests (RF), XGBoost, and decision trees (DT), as highlighted by \citeA{jain2020review} in their review of 300 different papers.

A common focus in the works exploring ML techniques has been the joint use or comparison between SVM and ANN models. For instance, \citeA{sayad2019predictive} combined in a single pipeline SVM and ANN to forecast wildfires using a dataset based on the NDVI, land surface temperature (LST), and thermal anomalies from the Moderate Resolution Imaging Spectroradiometer (MODIS), and three characteristics relevant to the status of crops. The findings demonstrated high fire occurrence prediction accuracy (ANNs: 98.32\%; SVM: 97.48\%), defining it as the ratio of correctly predicted instances to the total instances. The study integrated NDVI, LST, and thermal anomalies with ground truth labelled data. However, the researchers recommended integrating weather data due to its strong relationship with wildfire incidence, development, spread, and extinction. They especially highlighted utilising certain variables which drastically affect wildfire incidences and likelihood: air temperature, wind, and soil moisture \citeA{sayad2019predictive}. 

Another study included weather data, as suggested by \citeA{sayad2019predictive}. \citeA{sakr_efficient_2011} compared SVM and ANN for predicting the likelihood of a fire using cumulative precipitation and relative humidity for several months of the year. SVM performed better in binary fire/no fire classification with the highest accuracy of 94.21\%, while ANN obtained a maximum of 91.2\%.

%\citeA{pang2022forest} also included weather variables and aimed to develop a forest fire prediction model for China using a dataset collected from satellite remote sensing monitoring over 14 years, from 2003 to 2016. Before modelling, they conducted feature importance analysis using RF, highlighting fire hotspots as the most crucial variable, followed by relative humidity, sunshine hours, and average temperature. The study employed ANN, radial basis function network (RBF), SVM, and RF. The results demonstrated that RF outperformed other models, achieving the highest accuracy of 89.2\% and an Area Under Curve (AUC) of 0.96, making it the top-performing model in this research.

\citeA{gholamnia2020comparisons} took a distinct approach and explored ML models for mapping wildfire susceptibility instead of predicting them. Using wildfire inventory data from global positioning systems (GPS) and MODIS fire occurrences, they considered sixteen conditioning factors, including meteorological parameters and a vegetation index. Among the eleven ML techniques compared, RF showed the highest accuracy of 88\%, followed by SVM at 79\%. The study is noteworthy for implementing the unsupervised self-organising maps (SOM) technique, which was however applied to a two-class data and has been adapted for use for classification purposes. The least accurate models were RBF, logistic regression, and SOM. Although not directly comparable to the subject of this paper, it serves as a valuable example of ML's potential in the wildfire susceptibility mapping domain.

Turning attention to different geographical contexts, \citeA{sulova2020exploratory} conducted research similar to this study in terms of the study area, focusing on predicting wildfire risk in Australia over six months. They used topography, vegetation, infrastructure, meteorology, and socioeconomic data. The study employed Naïve Bayes (NB), Classification and Regression Tree (CART), and RF models, with RF achieving the highest accuracy of 96\% and NB having the lowest accuracy of 64\%. 
%Similar endeavours extended to different regions; \citeA{rodrigues2014insight}  conducted a study to assess wildfires in Spain, covering the entire peninsula except for specific cities. They used one of Europe's oldest databases, the Spanish EGIF database. On the data set, comparisons were made between linear regression, SVM, RF, and Boosting Regression Trees (BRT). The outcomes revealed that Random Forest and BRT displayed the best performance, with AUC values of 0.746 and 0.730, respectively. 

\citeA{kondylatos2022wildfire} aimed to accurately forecast wildfire danger using deep learning
(DL) models such as LSTM and ConvLSTM and traditional models like RF and XGBoost. The research covered Greece, parts of
the Balkan peninsula, and western Turkey. They utilised various datasets, which were including daily weather data from ERA-5 Land and satellite variables from MODIS, such as NDVI and Land Surface Temperature.
Results from the study showed that LSTM and ConvLSTM achieved F1-scores greater than 0.8, outperforming RF and XGBoost in all metrics. ConvLSTM exhibited fewer false positives and higher precision, while LSTM had the lowest
number of false negatives and highest recall.

%\citeA{gholami2021there} employed a dataset of wildfire occurrences in three landscapes in India from 2003 to 2016 to predict wildfire risk. Emphasising the temporal aspect of the data, they employed several models like logistic regression, SVM, XGBoost, bagging ensemble of decision trees (BDT), gradient boosting (GRB) and RF. The findings revealed that XGBoost and BDT outperformed others regarding the F1-score. 

% \citeA{liang2019neural} focused on forecasting the scale of forest wildfires using weather variables such as temperature, precipitation, and snow on the ground. The study utilises time series data from Alberta, Canada. The authors introduced a novel definition of wildfire scale, incorporating fire duration and burned area size. Three neural network models, including backpropagation neural network (BPNN), recurrent neural network (RNN), and LSTM, are examined for their ability to predict wildfire scale using meteorological data. The LSTM model demonstrates the highest predictive capability, achieving an overall accuracy of 90.9\%. The study sheds light on the connection between meteorological factors and wildfire scale in Alberta's forests. The authors proposed that the LSTM model could be adapted for predicting wildfire scales in other regions, too but acknowledged challenges due to variations in meteorological and environmental factors.

\citeA{arpaci2014using} examined fire occurrence data in Austria's Tyrol area from 1993 to 2011, incorporating terrain,
vegetation, climate, and socioeconomic factors. Their research intended to locate fire-prone locations and predict the
geographical spread of fires. They used the Maximum Entropy (MaxEnt) and RF methods to identify the primary variables
influencing the spread of fires. Climate change and population density have become essential factors in the
susceptibility of a given area to wildfires. RF model was able to forecast the distribution of fire ignition with a
classification accuracy of between 75 and 78\%. Both MaxEnt and RF algorithms were usually equivalent when categorising
fire susceptibility levels, except for the extremely high susceptibility class, where MaxEnt identified 28 places
compared to 2 by RF \cite{arif2021role}.

\citeA{malik2021wildfire} conducted fire risk prediction in San Diego, California, using weather and fire history data,
including temperature, soil moisture, relative humidity, and wind speed. SVM, XGBoost, and RF achieved accuracies of
98.18\%, 98.23\%, and 97.54\%, respectively, while the F1-scores were 97\%, 98\%, and 91\%, respectively.

\subsection{Anomaly Detection in Wildfire Prediction}
Anomaly detection aims to identify anomalous patterns, often known as outliers or anomalies, that significantly deviate
from the rest of the data \cite{thudumu2020comprehensive}. Such a task can be performed using supervised or unsupervised learning techniques. In this study, an unsupervised learning version will be implemented. In extreme event scenarios, it is difficult to find appropriately labelled datasets because detecting anomalous observations requires effort and expertise.
Therefore, supervised learning based on labelled data may not be a viable and suitable solution in all anomaly
detection circumstances \cite{esmaeili2023anomaly}. This underscores the value and contribution of unsupervised anomaly detection. 

To the best of our knowledge, there is no     study implementing anomaly detection directly in wildfire prediction using unsupervised learning techniques. However, there exists a closely related study working in wildfire risk prediction using anomaly detection method, even though it is not an exact match with this research's topic. In their research, \citeA{salehi_dynamic_2016} set out to use weather data to better predict the risk of wildfires. They devised a new model named Context-Based Fire Risk (CBFR) that considers the changing patterns of weather over time using a context-based anomaly
detection method. They used ECMWF ERA-5 weather data from the Blue Mountains in Australia. To assess the performance of their model, the authors compare it to the commonly used McArthur FFDI method. They looked at AUC and the straight-up accuracy rate. Their CBFR model performed better on both counts. Specifically, CBFR had an AUC of 0.96 and an accuracy of 0.94, while the FFDI method scored 0.89 for AUC and 0.87 for accuracy.

\subsection{Autoencoders and Dimensionality Reduction }
The first autoencoder neural network was developed to reduce dimensionality \cite{masci2011stacked}, showing specific advantages over other types of traditional dimensionality reduction methods. For instance, since they are non-linear, autoencoders can generally outperform linear traditional methods in the context of dimensionality reduction \cite{bank2020autoencoders}.

Although there have been no previous examples of such use of autoencoders in the wildfire prediction domain, there is one study focusing on post-fire assessment \cite{coca2021anomaly}. The authors introduced a framework for burned area estimation using multispectral images. Relying on ESA's Sentinel-2 (S2) sun synchronous satellite imagery, they focused on a wildfire-impacted area in Portugal from 2017. The authors employed an autoencoder to extract deep features from the data. They used the autoencoder's capability to not only reduce dimensionality, but also extract deep features from input data. After attaining these features, one-class Support Vector Machine (OCSVM) \cite{shin2005one} is utilised for anomaly detection. The experiment processed both normal and abnormal patches from the dataset, achieving an accuracy of 0.71 at 7.9\% abnormality. 

\subsection{Autoencoders and Anomaly Detection}
\subsubsection{Anomaly Detection Based on Reconstruction Error}
Using autoencoders to detect anomalies using the reconstruction error is a technique well applied in several
areas. The idea is as follows: a trained autoencoder would learn the latent subspace based on only normal
(non-anomalous) instances. Once trained, the input data is reconstructed, and reconstruction errors, i.e., the
difference between the original and reconstructed data, are calculated. As a result, while normal instances have low reconstruction errors, anomalies have high reconstruction errors \cite{bank2020autoencoders}.

To the best of our knowledge, no existing literature has explored anomaly detection using only deep autoencoders
and reconstruction error in the context of wildfire prediction. The forthcoming referenced studies, while
they are from varying domains, share a methodology closely aligned with this research.

\citeA{cheng2022resnet} utilised a ResNet autoencoder to identify anomalies in radar data. Their model incorporated both
convolutional and LSTM layers. Convolutional layers captured features, while LSTM layers
assessed temporal dependencies in radar data. The model achieved an accuracy potential of up to 85\%. \citeA{esmaeili2023anomaly} trained deep autoencoders on normal data to spot anomalies in a similar domain. Using autoencoder-based
prediction models on electrochemical aptasensor recordings with various signal lengths, they used reconstruction errors
to set up a threshold for assigning anomalies and used kernel density estimation. They implemented vanilla, ULSTM, and
Bi-LSTM autoencoder. The integrated ULSTM-vanilla model achieved about 80\% accuracy for longer signal datasets, 65\%
and 40\% for shorter ones.

\citeA{zhao2017spatio} introduced a unique model called Spatio-Temporal Autoencoder (STAE) in a different domain that learns
video representation and detects anomalies in the video by extracting features from spatial and temporal dimensions
using deep autoencoders. The utilisation of reconstruction error shares common points with the approach employed by
\citeA{hasan2016learning}. Once their model was trained, both studies produced a ``regularity
score" based on the reconstruction error, highlighting that video sequences with regular events had low
errors. In contrast, anomalous sequences had high errors, allowing them to spot anomalies. 

\subsubsection{Anomaly Detection Based on Clustering through Latent Features}
Clustering is organising unlabelled data into similar groupings known as clusters. A cluster is a group of data items
that are similar to one another but differ from other clusters \cite{serra2019unsupervised}. When labelled data is not
available, clustering offers a valuable alternative to supervised learning techniques.

Nevertheless, most traditional clustering algorithms are sensitive to the dimension of the input data, making them suffer from
 ``the curse of dimensionality” \cite{saxena2017review}, which makes them less effective in high-dimensional spaces than in
low-dimensional data \cite{aggarwal2013data}. In order to solve this problem, dimensionality reduction-based
clustering algorithms have been developed. These techniques translate data to lower dimensions utilising techniques
like PCA, kernel-based PCA, and autoencoders \cite{ma2022achieving}. For instance, \citeA{tian2014learning} described a simple method
that involved employing stacked autoencoders to acquire the non-linear embedding of the features (latent features) and
then feeding that information directly to k-means clustering. Similarly, \citeA{ma2022achieving} examined several models for various
clustering techniques on several datasets, including MNIST, HHAR, and REUTERS-10K. It has been noted that the Gaussian
Mixture Model (GMM) alone only achieved 47.8\% accuracy for the MNIST dataset. However, the accuracy rose to 79.34\%
when autoencoders and GMM were combined (AE+GMM), which might be a promising improvement. Like this, when AE and latent
characteristics were included, normalised mutual information (NMI) findings rose from 38.24\% to 81.41\%.

\section{Methodology}\label{method}
The code within this research has been developed using Python 3.10.10. The procedures related to
autoencoder modelling, including training, compilation, fine-tuning, and adaptation, were developed using TensorFlow 2.10.0 \cite{abadi_tensorflow:_2016}. In the context of clustering, the Scikit-learn 1.2.2 library was employed \cite{pedregosa_scikit-learn:_2018}.

\subsection{Data}

This paper focuses on the region of Australia. It includes seven states: New South Wales (NSW), Northern Territory (NT), Queensland (QL), South Australia (SA), Victoria (VI), Western Australia (WA), and Tasmania (TA). The dataset used is a combination of 2 datasets. The first dataset, released by IBM, is the historical weather and vegetation index. It was released for the Call for Code Spot Challenge for Wildfires competition, and is publicly available \cite{IBM}. It includes historical weather and vegetation index information from 2005 to 2021 as daily aggregated values. The second dataset, the Wildfires dataset is a component of the main dataset \cite{IBM} and it is only used to
evaluate the performances of the obtained model. Since we are following an unsupervised approach, the ``wildfire"
ground truth information obtained from the Wildfires dataset is not used during any training phase. Necessary
information regarding the datasets is summarised in Table \ref{tab_a}.

\begin{table}[!ht]
    \caption{Details of used datasets \cite{IBM}.}
    \centering
    \begin{tabular}{p{0.5\linewidth}  p{0.5\linewidth}}
    \toprule
        Dataset & Pre-processing details \\ \midrule
        Wildfires \cite{mcdl} & Spatial averaging into seven regions in Australia Aggregation by day from 2005 to 2021 Only high-confidence fires ($>$75\%) considered \\ %\hline
        Historical Weather - ERA5-Land hourly data \cite{era5land} & Raw ERA5 data was processed using IBM PAIRS Geoscope Temporal aggregation, unit conversion, and transformations are applied to each parameter. Spatial aggregation using physical area weights of each pixel \\ %\hline
        NDVI - MOD13Q1 MODIS/Terra Vegetation Indices 16-Day L3 Global 250m SIN Grid V006 \cite{ndvi}  & Spatial averaging into seven regions in Australia from 2005 to 2021 \\ \bottomrule
    \end{tabular}\label{tab_a}
\end{table}

The data presented is aggregated daily and encompasses estimations by NASA, having
undergone certain processing stages \cite{IBM}. 

\begin{figure}[!ht]
\noindent
\begin{center}
 \includegraphics[width=1.05\textwidth]{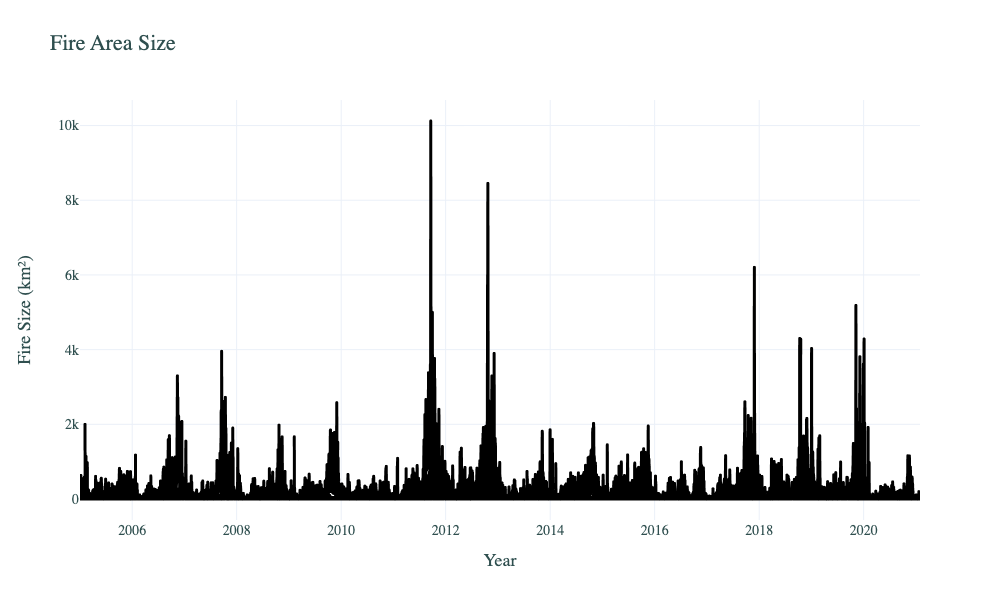} 
 \end{center}
\caption{Fire area size over the years for all regions.}
\label{fig_a1}
\end{figure}

Figure \ref{fig_a1} provides an overview of fire area size across all Australian regions from 2005 to 2021. A significant surge in fire incidents stands out in 2006 and 2007. Remarkably, there was another notable increase in active fire occurrences during 2011 and 2012, followed by a subsequent rise in wildfires during 2019 and 2020.

The two initial datasets (Weather and NDVI) were merged to be used as input data. Concerning the target variable, given that this is an unsupervised learning work, the actual wildfire ground truth is not utilised during the training and modelling phases. Nonetheless, to evaluate model performance and outcomes, a binary target variable labelled as``Fire" is generated from the Estimated Fire Area Size variable using the subsequent formula:

\begin{equation}
\mathrm{Fire}=\left\{\begin{matrix}1,\mathrm{Estimated\ Fire\ Area\ }{\neq}0\\~0,\mathrm{Estimated\ Fire\ Area\ }=0\end{matrix}\right.
\end{equation}

\begin{table}[!ht]
\caption{Details of features.}
    \centering
    \begin{tabular}{p{0.1\linewidth}  p{0.3\linewidth} p{0.5\linewidth}}
    \toprule
        Dataset  & Features & Range and Unit \\ \midrule
        Wildfires  & Date  & 2005-01-01 - 2021-01-23 \\ %\hline
        ~ & Region & [NSW, NT, QL, SA, VI, WA, TA] \\ %\hline
        ~ & Estimated Fire Area Size  & 0 $\leq$ Estimated Fire Area Size (km$^2$)$<$10120.94 \\ %\hline
        ~ & (Estimated Fire Area Size is converted to a binary variable) Fire  & [0,1] \\ %\hline
        Historical Weather  & Date & 2005-01-01 - 2021-01-23 \\ %\hline
        ~ & Region & [NSW, NT, QL, SA, VI, WA, TA] \\ %\hline
        ~ & Minimum, maximum, mean, and variance of each of these parameters: Precipitation &    0 $\leq$ Precipitation (mm/day) $\leq$ 509.83 \\ %\hline
        ~ & Relative humidity & 0 $\leq$ Relative humidity (mm/day) $\leq$ 509.83 \\ %\hline
        ~ & Soil water content & 0 $\leq$ Soil water content (m$^3$ m$^{-3}$) $\leq$ 0.52 \\ %\hline
        ~ & Solar radiation & 0.41 $\leq$   Solar radiation (MJ/day) $\leq$ 35.69 \\ %\hline
        ~ & Temperature  & -5.05 $\leq$   Temperature (°C) $\leq$ 41.73 \\ %\hline
        ~ & Wind speed & 0.25 $\leq$   Wind speed (m/s) $\leq$ 24.27 \\ %\hline
        NDVI  & Date & 2005-01-01 - 2021-01-23 \\ %\hline
        ~ & Region & [NSW, NT, QL, SA, VI, WA, TA] \\ %\hline
        ~ & Minimum, maximum, mean, and variance of: NDVI &   0 $\leq$ NDVI (index) $\leq$ 1 \\ \bottomrule
    \end{tabular}\label{tab_b}
\end{table}

Table \ref{tab_b} provides an overall list of all features and their range and units. In the final version of the dataset, there are 26,677 instances of wildfires and 14,299 instances of non-wildfires. This
imbalance arises because the dataset is aggregated daily. If even a single wildfire occurs on a given day, that day
is labelled as a  ``wildfire" day.

\subsubsection{Feature Importance }
\begin{figure}[!ht]
\noindent
\begin{center}
 \includegraphics[width=1.0\textwidth]{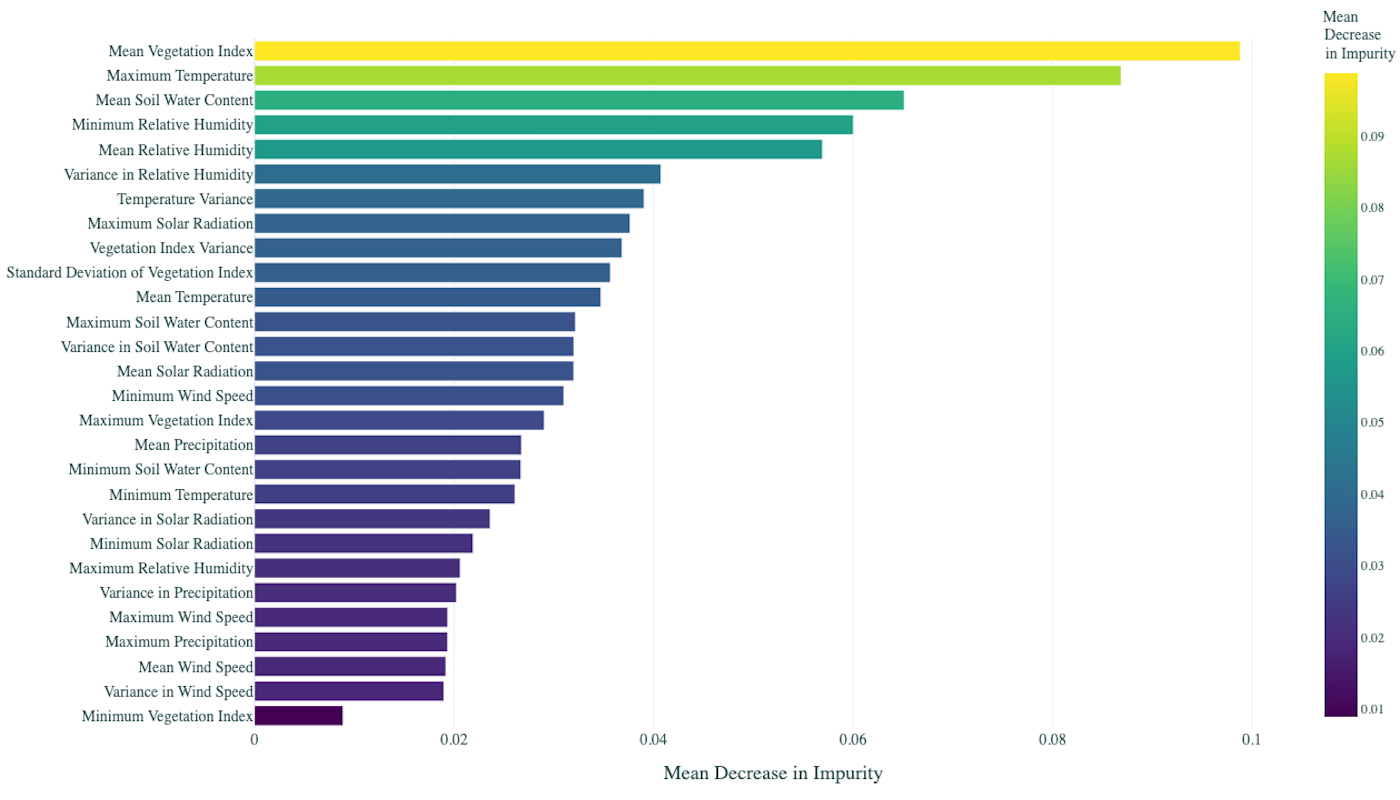} 
 \end{center}
\caption{Feature importance values generated by RF.}
\label{fig_a}
\end{figure}

Given that certain variables might be interrelated, and some may not provide significant value, RF models inherently
select a subset of significant variables for classification, instead of using all variables, and thus enhancing the classification performance of high-dimensional data. This
feature selection can be visualized through “Gini importance”, also known as mean decrease in impurity (MDI), which
ranks the relevance of features \cite{sutera_importance_2021}. RF was employed to get a feature importance plot with MDI values, as
seen in Figure \ref{fig_a}. Parameters used for the RF fit can be found in Table \ref{tab_b2}.

\begin{table}[!ht]
\caption{Parameters used in the RF fit.}
    \centering
    \begin{tabular}{ll}
    \toprule
        Parameter & Value \\ \midrule
        Number of estimators & 100  \\ %\hline
        Criterion & “Gini”  \\ %\hline
        Min samples split & 2  \\ %\hline
        Min samples leaf & 2   \\ \bottomrule
    \end{tabular}\label{tab_b2}
\end{table}

Figure \ref{fig_a} provides evidence that several variables, such as the Vegetation Index Min, Wind Speed Mean or Wind Speed Variance, contribute minimally to the model. This observation is particularly pronounced in the RF, highlighting
numerous less impactful features. 

In the deep autoencoder modelling part, two datasets will be analysed:

\begin{enumerate}
\item \textbf{Dataset 1} includes all 28 variables without any exclusions. 
\item \textbf{Dataset 2} is composed of 17 features. Based on a close look at Figure \ref{fig_a}, it is seen that there is a significant
decrease in MDI after the Vegetation Index Max value. Yet, since many studies emphasize the significance of the Temperature
variable in this field, Temperature Min is retained as well. Consequently, any variable with an MDI value below
Temperature Min was manually removed for experimental purposes. 
\end{enumerate}
Comparison of these two dataset results will allow us to assess the models' performance concerning
variable contribution and dimensional variance. The complete set of features for both data versions can be found in
Table \ref{tab_c}. `Region' and `Date' features are omitted as they do not offer any insights into
the model.

\begin{table}[!ht]
\caption{Features for two versions of the dataset.}
    \centering
    \begin{tabular}{p{0.3\linewidth}  p{0.3\linewidth}}
    \toprule
        Dataset 1 &  Dataset 2 \\ \midrule
        Precipitation\_Max & Vegetation\_index\_Mean \\ %\hline
        Precipitation\_Mean & Temperature\_Max \\ %\hline
        Precipitation\_Variance & SoilWaterContent\_Mean \\ %\hline
        RelativeHumidity\_Max & RelativeHumidity\_Min \\ %\hline
        RelativeHumidity\_Mean & RelativeHumidity\_Mean \\ %\hline
        RelativeHumidity\_Min & RelativeHumidity\_Variance \\ %\hline
        RelativeHumidity\_Variance & Temperature\_Variance \\ %\hline
        SoilWaterContent\_Max & SolarRadiation\_Max \\ %\hline
        SoilWaterContent\_Mean & Vegetation\_index\_Std \\ %\hline
        SoilWaterContent\_Min & Vegetation\_index\_Variance \\ %\hline
        SoilWaterContent\_Variance & Temperature\_Mean \\ %\hline
        SolarRadiation\_Max & SoilWaterContent\_Max \\ %\hline
        SolarRadiation\_Mean & SoilWaterContent\_Variance \\ %\hline
        SolarRadiation\_Min & SolarRadiation\_Mean \\ %\hline
        SolarRadiation\_Variance & WindSpeed\_Min \\ %\hline
        Temperature\_Max & Vegetation\_index\_max \\ %\hline
        Temperature\_Mean & Temperature\_Min \\ %\hline
        Temperature\_Min & ~ \\ %\hline
        Temperature\_Variance & ~ \\ %\hline
          WindSpeed\_Max & ~ \\ %\hline
        WindSpeed\_Mean & ~ \\ %\hline
        WindSpeed\_Min & ~ \\ %\hline
        WindSpeed\_Variance & ~ \\ %\hline
        Vegetation\_index\_Mean & ~ \\ %\hline
        Vegetation\_index\_Max & ~ \\ %\hline
        Vegetation\_index\_Min & ~ \\ %\hline
        Vegetation\_index\_Std & ~ \\ %\hline
        Vegetation\_index\_Variance & ~ \\ \bottomrule
    \end{tabular}\label{tab_c}
\end{table}

\subsubsection{Data Split}
The data is split into three parts: training, validation, and testing. For training, data with only normal (non-wildfire) cases are
used. A mix of data from both classes is needed for the validation and test parts. The original dataset has 14,299 non-wildfire (NW) cases (35\%) and 26,677 wildfire (W) cases (65\%). This
imbalance within the classes arises due to the daily aggregation of data. If even a single minor wildfire is reported
on a given day, that day is labelled as a  ``wildfire", increasing the wildfire count.

The autoencoder is trained solely on non-wildfire data. Therefore, a substantial portion of NW from the original
dataset (12,869 cases) is allocated to the train data, ensuring maximum use of the non-wildfire information present. The remaining NW cases are evenly divided between the test and validation datasets, with each receiving 715 NW
cases. For test and validation datasets, wildfire cases are initially split into 13,338 and 13,339, respectively.
However, given that the train data consisted of a total of 12,869 instances, these numbers seemed disproportionate. To
address this, 1,000 wildfire cases are randomly dropped for both the test and validation datasets. Thus, each test and validation set consists of 1,000 wildfire cases with 715 non-wildfire cases, resulting in
datasets with (W=1,000, NW=715).

This process ensures that the test and validation dataset sizes are not excessively larger than the train data due to
the class imbalance. Consequently, the distribution for training, testing, and validation sets is roughly 80\%, 10\%, and
10\%. Still, it is worth noting that, since some instances had to be dropped due to the imbalance in the original dataset,
some information in the original data was not fully utilized during modelling, which could be listed as a limitation.
This whole procedure is applied separately for both Dataset 1 and Dataset 2. Min-max scaling was employed to ensure that all features in the dataset have a uniform scale, allowing them to contribute equally to the model's performance. Each dataset - train, validation, and test, was scaled based on its own statistics. Therefore, it was ensured that data leakage did not occur, preserving the integrity of the evaluation. Through this scaling procedure, it is ensured that every feature within the datasets is within the range of 0 to 1.

\subsection{Methods }

In this study, we focus on two main unsupervised approaches: using a deep autoencoder to obtain latent features for clustering models and using a deep autoencoder to reconstruct input data and identify anomalies through reconstruction errors. These strategies have been discussed extensively in the Related Work section, where their applications and effectiveness in different contexts were reviewed in detail. The main aim here is to fill the research gap in unsupervised learning for wildfire prediction using
unlabelled data. Deep autoencoders are utilised to detect anomalies in data, classifying them as
 ``wildfire" or  ``non-wildfire." The model is trained on
non-anomaly instances. Two strategies are explored:

\begin{enumerate}
\item Using the deep autoencoder for training, then reconstructing input data in lower dimensions and applying clustering
methods to these latent features to differentiate wildfires.
\item Using the deep autoencoder to reconstruct input data, where high reconstruction errors signify anomalies. Here,
both Fully Convolutional (FC) and Long-Short-Term-Memory (LSTM) autoencoders are employed, with the latter being adept at capturing temporal patterns in time-series data, a feature that FC architectures lack.
\end{enumerate}

In the following sections, we start by explaining what a deep autoencoder is in Section \ref{sec:deep_autoencoder}, including its variants such as Fully Connected (FC) and LSTM Autoencoders. Then, in Sections \ref{anom_clust} and \ref{anom_error}, we detail the two main methods utilized in this paper that employ deep autoencoders for anomaly detection.

\subsubsection{Deep Autoencoders}\label{sec:deep_autoencoder}
\begin{figure}[!ht]
\noindent
\begin{center}
 \includegraphics[width=1.0\textwidth]{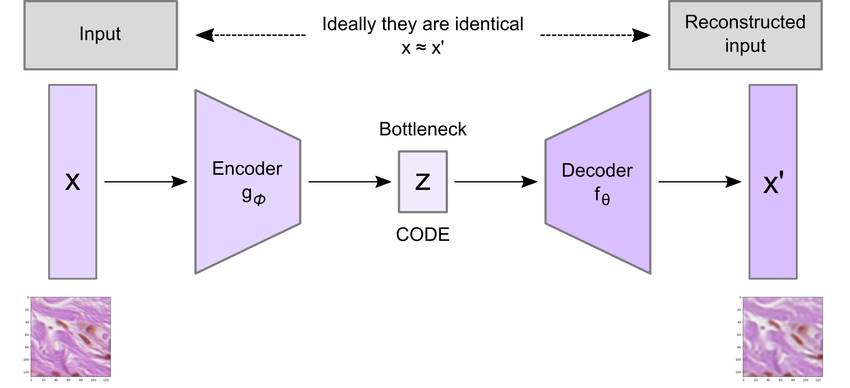}
 \end{center} 
\caption{Architecture of a sample autoencoder \cite{kucharski_semi-supervised_2020}}
\label{fig_b}
\end{figure}

An autoencoder is a unique variant of feed-forward neural networks aiming to have the output closely mirror the input. Figure \ref{fig_b} shows that both the input and output layers have matching dimensions. An autoencoder consists of two main parts: an
encoder and a decoder as provided in Figure \ref{fig_b}.

The encoder section comprises one or multiple hidden layers, in this case, FC or LSTM layers, that compress the input
into a condensed representation. Conversely, the decoder uses its hidden layers, which can also be FC or LSTM layers,
to reconstruct this condensed representation back to the original input's dimensionality \cite{kucharski_semi-supervised_2020}. The
unsupervised training process optimises these hidden layers to minimise the difference between the reconstructed output
and the original input. This study will explore autoencoders with FC and LSTM hidden layers to comprehensively
understand their performance and capabilities.
   
\paragraph{LSTM Autoencoder.}
As mentioned, a deep autoencoder can include hidden LSTM layers in its encoder and decoder components. Using LSTM layers
provides the advantage of capturing temporal information, if any, in the time-series data. To utilise the LSTM autoencoder, the dataset must be reshaped into a format compatible with what the LSTM requires.
Thus, for the training, testing, and validation datasets, sequences of a 10-day length were prepared. Table \ref{tab_d} showcases the precise shape details of the inputs prior to feeding them into the LSTM autoencoder. Table \ref{tab_e} provides detailed
information on the final model's hyperparameter tuning details and the library used for this.

\begin{table}[!ht]
\caption{Shape of the datasets to be used in the LSTM autoencoder.}
    \centering
    \begin{tabular}{p{0.22\linewidth}  p{0.23\linewidth} p{0.23\linewidth} p{0.23\linewidth}}     \toprule
        ~ & Train (batch, timesteps, feature) & Test (batch, timesteps, feature) & Validation (batch, timesteps, feature) \\ \midrule
        Shape (Dataset 1)  & (1286, 10, 28) & (171, 10, 28) & (171, 10, 28) \\ %\hline
        Shape (Dataset 2)  & (1286, 10, 17) & (171, 10, 17) & (171, 10, 17) \\ \bottomrule
    \end{tabular}\label{tab_d}
\end{table}

\begin{table}[!ht]
\caption{Hyperparameter tuning details for LSTM autoencoder.}
    \centering
    \begin{tabular}{p{0.3\linewidth}  p{0.45\linewidth} p{0.25\linewidth}}     \toprule
        Parameter \& Library & Range &  Final Choice \\ \midrule
        Number of layers (tf.keras.layers.LSTM)  & [2, 3, 4, 5, 6, 7, 8] & 5 \\ %\hline
        Number of units (tf.keras.layers.LSTM)  & [1024, 512, 256, 128, 64, 32, 16, 8] & [256, 128, 64, 32,16] \\ %\hline
        Activation function (tf.keras.layers.LSTM) & [ReLU, Sigmoid, Tanh, Softmax] & Tanh \\ %\hline
        Optimizer (tf.keras.optimizers) & [Adam, SGD, RMSProp] & Adam \\ %\hline
        Learning rate scheduler (tfa.optimizers) & Exponential Decay, Polynomial Decay, Inverse Time Decay, Cosine Decay, Cyclical Learning Rate, None & None  \\ %\hline
        Batch size  & [16, 32, 64, 128, 256, 512] & 32 \\ \bottomrule
    \end{tabular}\label{tab_e}
\end{table}

\begin{figure}[!ht]
\noindent
\begin{center}
 \includegraphics[width=0.3\textwidth]{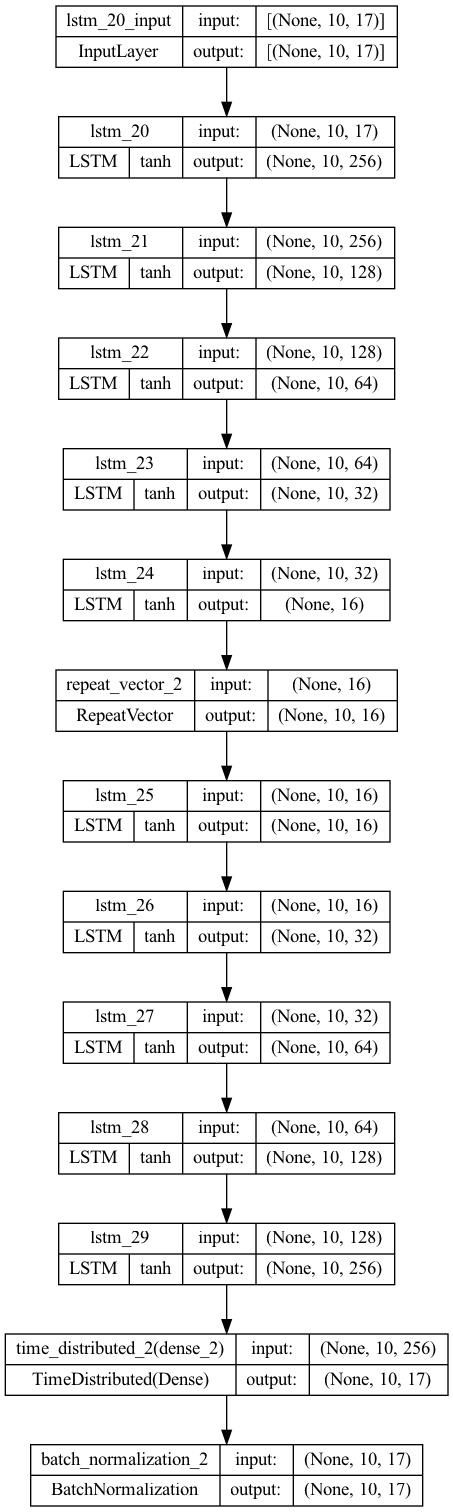} 
 \end{center}
\caption{Final LSTM autoencoder architecture.}
\label{fig_c}
\end{figure}

Figure \ref{fig_c} represents the LSTM autoencoder architecture. The design uses a stacked LSTM structure, where each layer feeds
its output into the subsequent layer as its input. Since it is an autoencoder, a mirrored architecture could be
observed in the layers.

\begin{figure}[!ht]
\noindent
\begin{center}
 \includegraphics[width=0.3\textwidth]{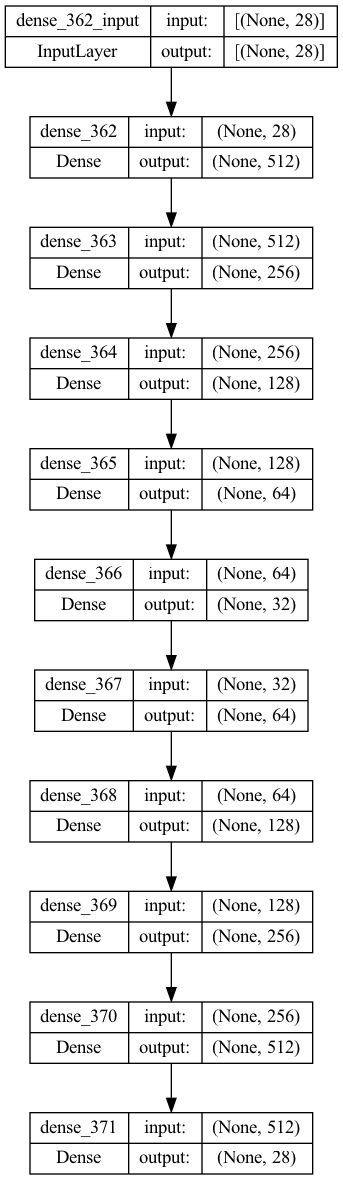} 
 \end{center}
\caption{Final FC autoencoder architecture.}
\label{fig_d}
\end{figure}

\paragraph{Fully Connected (FC) Autoencoder}
Like the LSTM autoencoder, the FC autoencoder incorporates hidden, fully connected layers in both its encoder and
decoder parts. The hyperparameter tuning details for the FC
autoencoder are detailed in Table \ref{tab_f}, which shows the final model setup and the parameters explored.  Like the LSTM autoencoder, the FC autoencoder also establishes mirrored architectures, formed with FC layers. Figure \ref{fig_d} represents the FC autoencoder architecture.

\begin{table}[!ht]
\caption{Hyperparameter tuning details for FC autoencoder}
    \centering
    \begin{tabular}{p{0.3\linewidth}  p{0.45\linewidth} p{0.25\linewidth}}     \toprule
        Parameter & Range & Final Choice \\ \midrule
        Number of layers (tf.keras.layers.LSTM)  & [2, 3, 4, 5, 6, 7, 8] & 5 \\ %\hline
        Number of units (tf.keras.layers.LSTM)  & [1024, 512, 256, 128, 64, 32, 16, 8] & [512, 256, 128, 64,32] \\ %\hline
        Activation function (tf.keras.layers.LSTM) & [ReLU, Sigmoid, Tanh, Softmax] & ReLU \\ %\hline
        Optimizer (tf.keras.optimizers) & [Adam, SGD, RMSProp] & Adam \\ %\hline
        Learning rate scheduler (tfa.optimizers) & Exponential Decay, Polynomial Decay, Inverse Time Decay, Cosine Decay, Cyclical Learning Rate, None & Cyclical Learning Rate \\ %\hline
        Batch size  & [16, 32, 64, 128, 256, 512] & 128 \\ \bottomrule
    \end{tabular}\label{tab_f}
\end{table}

\subsubsection{Anomaly Detection with Deep Autoencoders}

This section delves into the anomaly detection methodology utilising the deep autoencoder models discussed in the
preceding sections. 

The choice of the two detection strategies is informed by their proven efficacy in the domain of anomaly detection, as highlighted in Section \ref{related_work}. The first approach, using latent features obtained from a deep autoencoder for clustering models, leverages the ability of autoencoders to reduce dimensionality and extract significant features from the data. This method has shown promise in various applications, including environmental monitoring and anomaly detection, as discussed by authors such as \citeA{tian2014learning} and 
\citeA{ma2022achieving}. The clustering models used—isolation forest, local outlier factor (LOF), and one-class SVM—are well-established techniques for identifying outliers in high-dimensional data \cite{liu2012isolation,breunig2000lof,scholkopf1999support}.

The second approach, using reconstruction errors from a deep autoencoder to identify anomalies, exploits the inherent capability of autoencoders to learn the normal patterns in the data. When the autoencoder reconstructs data points, anomalies typically result in higher reconstruction errors, which can be flagged as outliers. This method has been successfully applied in various fields, including video surveillance and medical diagnostics, as demonstrated by researchers such as \citeA{cheng2022resnet} and \citeA{esmaeili2023anomaly}.

Implementation details of these two primary methods are explained in more detail in the following sections.

\paragraph{Anomaly Detection Based on Clustering through Latent Features.}\label{anom_clust}

In this case, the encoder segment of the FC autoencoder is employed to extract the latent features from the train and
test datasets. These features are compressed from a 28-dimensional space down to an 8-dimensional one. An FC autoencoder is used, pre-trained on the train data (which consists of only non-wildfire cases). Specifically, to obtain latent features, the encoder component of the FC autoencoder is employed.
The data dimension for both is reduced from 28 to 8. Then, latent train data is used to fit the cluster model. Three clustering techniques are adopted
separately here: isolation forest, LOF, and one-class SVM. Finally, the prediction of the latent test data is done using the clustering model, and anomalies are identified as a result. Each clustering model has multiple parameters that require tuning. 

Each method has unique characteristics that make it suitable for specific aspects of anomaly detection in high-dimensional data.
The isolation forest method, which isolates anomalies rather than profiling normal instances, operates by constructing binary trees through random selection of features and split values, creating shorter paths for anomalies due to their rarity in the dataset, offering rapid anomaly detection efficiency \cite{liu2012isolation}. LOF measures the local density deviation of a data point relative to its neighbors, identifying anomalies in areas of varying density \cite{breunig2000lof}. Lastly, one-class SVM  is designed for scenarios where only the `normal' class is present during training; it defines a decision boundary around this class, with deviations from this boundary treated as anomalies, making it commonly suitable for the preliminary phases of environmental anomaly detection of the unsupervised learning scenarios \cite{scholkopf1999support}.  The incorporation of isolation forest, LOF, and one-class SVM in this study is designed to facilitate a comprehensive comparative analysis. By evaluating each method's performance, the study aims to provide essential insights into their comparative strengths and weaknesses, thereby informing future research and practical applications in the field. To achieve this, all three methods are implemented concurrently, allowing for a direct and comprehensive comparison of their results.

Table \ref{tab_g} provides the parameters adjusted for each
clustering model and the final choice based on the evaluation results. Only Dataset 1 is utilised for this approach.

\begin{table}[!ht]
\caption{Hyperparameter tuning details for clustering models.}
    \centering
    \begin{tabular}{p{0.2\linewidth}  p{0.25\linewidth} p{0.4\linewidth} p{0.1\linewidth}}     \toprule
        ~ & Parameter & Range  & Final Choice \\ \midrule
        Isolation forest & Number of estimators & [50, 100, 200] & 200 \\ %\hline
        ~ & Max samples & [0.5, 0.7, 0.9] & 0.9 \\ %\hline
        ~ & Contamination & [`auto', 0.01, 0.02, 0.05, 0.1, 0.2,0.3, 0.4, 0.5] & 0.5 \\ %\hline
        LOF & Number of neighbours & [2, 3, 4, 5, 6, 7, 8, 10, 12, 15, 20] & 8 \\ %\hline
        ~ & Contamination & [`auto', 0.01, 0.02, 0.05, 0.1, 0.2,0.3, 0.4, 0.5] & 0.5 \\ %\hline
        ~ & Distance metric & [Euclidean, Manhattan, Minkowski] & Manhattan \\ %\hline
        One class SVM & Kernel & [Linear, Polynomial, RBF, Sigmoid] & Linear \\ %\hline
        ~ & Nu value & [0.01, 0.05, 0.1, 0.2, 0.3,0.4,0.5,0.6]    & 0.6 \\ \bottomrule
    \end{tabular}\label{tab_g}
\end{table}

\paragraph{Anomaly Detection Based on Reconstruction Error.}\label{anom_error}
In this approach, both FC and LSTM autoencoders are employed independently, though the procedure for each remains the same. The hyperparameter tuning details for both FC and LSTM autoencoders are detailed in Table \ref{tab_gg}. The selected LSTM model is presented in table \ref{tab_gg2}. This model uses a batch size of 32, Tanh activation functions for both the
encoder, decoder, and Adam optimizer with LSTM layers [256, 128, 64, 32,16].

\begin{table}[!ht]
\caption{FC and LSTM autoencoder final hyperparameter tuning range.}
    \centering
    \begin{tabular}{p{0.5\linewidth}  p{0.5\linewidth}}     \toprule
        Parameter \& Library & Range \\ \midrule
        Activation function
(tf.keras.layers.LSTM) & [ReLU, Sigmoid, Tanh, Softmax] \\ %\hline
        Optimizer
(tf.keras.optimizers) & [Adam, SGD, RMSProp] \\ %\hline
        Learning rate scheduler
(tfa.optimizers) & Exponential Decay, Polynomial Decay, Inverse Time Decay, Cosine Decay, Cyclical Learning Rate, None \\ %\hline
        Batch size & [16, 32, 64, 128, 256, 512] \\ \bottomrule
    \end{tabular}\label{tab_gg}
\end{table}

\begin{table}[!ht]
\caption{LSTM autoencoder model selected after hyperparameter
tuning on the test data.}
    \centering
    \begin{tabular}{p{0.23\linewidth} p{0.23\linewidth} p{0.23\linewidth} p{0.23\linewidth} }
    \toprule
        Batch Size & Encoder Activation Function & Decoder Activation Function & Optimizer  \\ \midrule
        32 & Tanh & Tanh & Adam \\ \bottomrule
    \end{tabular}\label{tab_gg2}
\end{table}

Hyperparameter tuning was exclusively conducted for Dataset 1. Consequently, the same architecture of the chosen final
model was also used when training Dataset 2. For the FC case, two models emerge as
strong candidates for the final selection. These top-tier models will be referred to as Model A and Model B and are presented in table \ref{tab_ii}. Having the highest F1 scores, they indicate a balance
between precision and recall and a decent overall classifier performance. It is decided to assess both and compare their results, before
finalizing the choice. Each model is trained separately on both Dataset 1 and Dataset 2 (on the test data).

\begin{table}[!ht]
\caption{FC autoencoder models selected after hyperparameter
tuning on the test data.}
    \centering
    \begin{tabular}{p{0.12\linewidth} p{0.12\linewidth} p{0.12\linewidth} p{0.12\linewidth} p{0.3\linewidth} }
    \toprule
        Model & Activation Function & Batch Size & Optimizer & Learning Rate Scheduler \\ \midrule
         Model A & ReLU  & 128 & Adam & Cyclical Learning Rate \\ %\hline
         Model B & ReLU  & 32 & RMSProp & None \\ \bottomrule
    \end{tabular}\label{tab_ii}
\end{table}

\subparagraph{Training Process.}
The training dataset only consists of non-wildfire instances, while both the test and validation datasets contain a mix of
cases. After obtaining the best combination of parameters for the model and
finalizing the autoencoder model, training is performed. During the training, validation data is used to observe the
validation loss and errors for each iteration of the training process.

Initially, the train data is
input into the FC/LSTM autoencoder, which has been previously trained on this data. The autoencoder then generates the
reconstructed train data as its output. The reconstruction error, represented as the mean squared logarithmic error (MSLE), quantifies the deviation
between the original and the reconstructed train data. Therefore, subsequently, MSLETrain values are calculated to
calculate the difference between the original train data and reconstructed train data.  Using the MSLETrain values, a threshold is set up based on the formula provided below, here termed as
ThresholdTrain.

\begin{equation}
\mathrm{Threshold}_{\mathrm{Train}}=\mu _{\mathrm{MSLE}(\mathrm{Train})}+2\sigma _{\mathrm{MSLE}(\mathrm{Train})}
\end{equation}

For the procedure for test data reconstruction and anomaly detection, similarly, the test data is
input into the same FC/LSTM autoencoder. The autoencoder then generates the reconstructed test data as its output. Subsequently, MSLETest values are calculated to calculate the difference between the original test data and
reconstructed test data. Finally, in the classification phase, MSLETest values are compared with ThresholdTrain to assign anomalies. Basically,
if MSLETest is higher than ThresholdTrain, then the data point is categorized as a wildfire, denoted by “1”. If not, it
categorized as a non-wildfire, denoted by “0”, with the following formula:

\begin{equation}
\mathrm{Fire}_{\mathrm{Test}}=\left\{\begin{matrix}1,\mathrm{Threshold}_{\mathrm{Train}}~{\leq}~\mathrm{MSLE}_{\mathrm{Test}}\\~0,\mathrm{Threshold}_{\mathrm{Train}}>\mathrm{MSLE}_{\mathrm{Test}}\end{matrix}\right.
\end{equation}

\paragraph{Performance Metrics}
For both approaches, after getting the anomaly detection/classification results based on test data, the performance of
the model on unseen data needs to be evaluated. For this, the model's performance is evaluated against
five metrics: Accuracy, precision, recall, F1-score, and Matthew's coefficient (MCC).
The following equations present formulas for calculating these metrics, where TP = True Positive, TN = True Negative,
FP= False Positive, and FN = False Negative:

\begin{equation}
\mathrm{Accuracy}=\frac{\mathit{TP}+\mathit{TN}}{P+N}
\end{equation}
\begin{equation}
\mathrm{Precision}=\frac{\mathit{TP}}{\mathit{TP}+\mathit{FP}}
\end{equation}
\begin{equation}
\mathrm{Recall}=\frac{\mathit{TP}}{\mathit{TP}+\mathit{FN}}
\end{equation}
\begin{equation}
\mathrm{F1}=2\frac{(\mathrm{Precision}\times\mathrm{Recall})}{(\mathrm{Precision}+\mathrm{Recall})}
\end{equation}
\begin{equation}
\mathrm{Matthew's\ coefficient\ (MCC)}=\frac{\mathrm{TP}\times\mathrm{TN}-\mathrm{FP}\times\mathrm{FN}}{\sqrt{(\mathrm{TP}+\mathrm{FP})(\mathrm{TP}+\mathrm{FN})(\mathrm{TN}+\mathrm{FP})(\mathrm{TN}+\mathrm{FN})}}
\end{equation}

A confusion matrix for each model \cite{demir_deep_2022}, based on test data, was also produced. Lastly, a ROC curve plot and AUC are plotted and calculated. A ROC curve plots the TP rate in
comparison to the FP rate. These values shift relative to one another. Essentially, when the TP rate peaks, the FP rate
is at its minimum, and vice versa \cite{maxion2004proper}. AUC is a key metric for assessing a binary classifier's performance. It ranges from 0.5 (equivalent to random guessing)
to 1.0 (a flawless classifier). AUC evaluates the effectiveness of score classifiers by considering all potential
classification thresholds. \cite{melo2013area}.

\section{Results}\label{results}

\subsection{Anomaly Detection Based on Clustering through Latent Features}

In this section, two different models and their results will be presented for each cluster model type (Isolation
Forest, LOF, and one-class SVM). While the first
one is the best model according to the hyperparameter tuning, the other one will have different parameters for the sake
of the comparison of the parameters and performance.

\subsubsection{Isolation Forest}

\begin{table}[!ht]
\caption{Parameter information and results for two models of isolation forest.}
    \centering
    \begin{tabular}{p{0.09\linewidth}  p{0.09\linewidth} p{0.09\linewidth} p{0.09\linewidth} p{0.09\linewidth} p{0.09\linewidth} p{0.09\linewidth} p{0.09\linewidth} p{0.09\linewidth}}     \toprule
        Model & Number of estimators & Max samples & Contami-nation & Accuracy & Precision & Recall & F1-score & MCC \\ \midrule
        Model (A) & 100 & 0.9 & 0.5 & 0.612 & 0.664 & 0.675 & 0.67 &  0.242 \\ %\hline
        Model (B) & 100 & 0.9 & 0.2 & 0.536 & 0.717 & 0.337 & 0.459 &  0.264 \\ \bottomrule
    \end{tabular}\label{tab_h}
\end{table}

\begin{figure}[!ht]
\noindent
\begin{center}
 \includegraphics[width=1.0\textwidth]{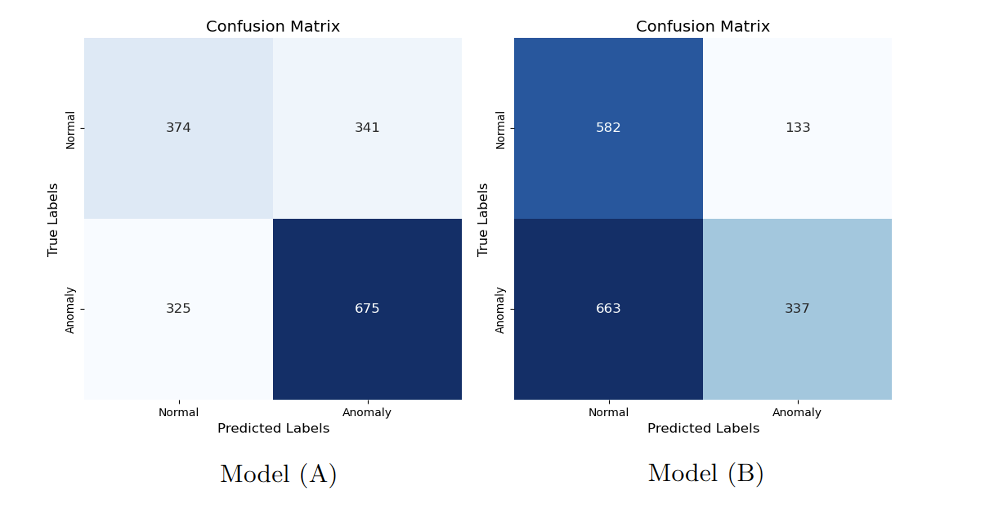} 
 \end{center}
\caption{Confusion matrix comparison for Model (A) and (B) for isolation forest.}
\label{fig_e}
\end{figure}

Based on Table \ref{tab_h}, Model (A) assumes half of the data is anomalous (0.5), while Model (B) only assumes 0.2 of the data is anomalous
due to the contamination parameter. This discrepancy leads to a significant difference in their performance. 

Figure \ref{fig_e} illustrates Model (A) has a better balance between precision and recall, resulting in a higher F1-score of 67\%. In contrast,
Model (B) has a higher precision of 71.7\%, suggesting that among the anomalies it detected, a greater percentage were
true anomalies. However, its recall is considerably low at 33.7\%, which means it failed to capture a significant proportion
(around 66\%) of the true anomalies present in the dataset.

\subsubsection{LOF}

\begin{table}[!ht]
\caption{Parameter information and results for two models of LOF}
    \centering
    \begin{tabular}{p{0.1\linewidth}  p{0.1\linewidth} p{0.1\linewidth} p{0.1\linewidth} p{0.1\linewidth} p{0.1\linewidth} p{0.1\linewidth} p{0.1\linewidth} p{0.1\linewidth}}     \toprule
        Model & Number of neighbours & Contami-nation & Distance Metric & Accuracy & Precision & Recall & F1-score & MCC \\ \midrule
        Model (C) & 20 & 0.5 & Manhattan & 0.65 & 0.677 & 0.766 & 0.719 &  0.329 \\ %\hline
        Model (D) & 12 & 0.3 & Manhattan & 0.631 & 0.72 & 0.6 & 0.655 & 0.361 \\ \bottomrule
    \end{tabular}\label{tab_h2}
\end{table}

\begin{figure}[!ht]
\noindent
\begin{center}
 \includegraphics[width=1.0\textwidth]{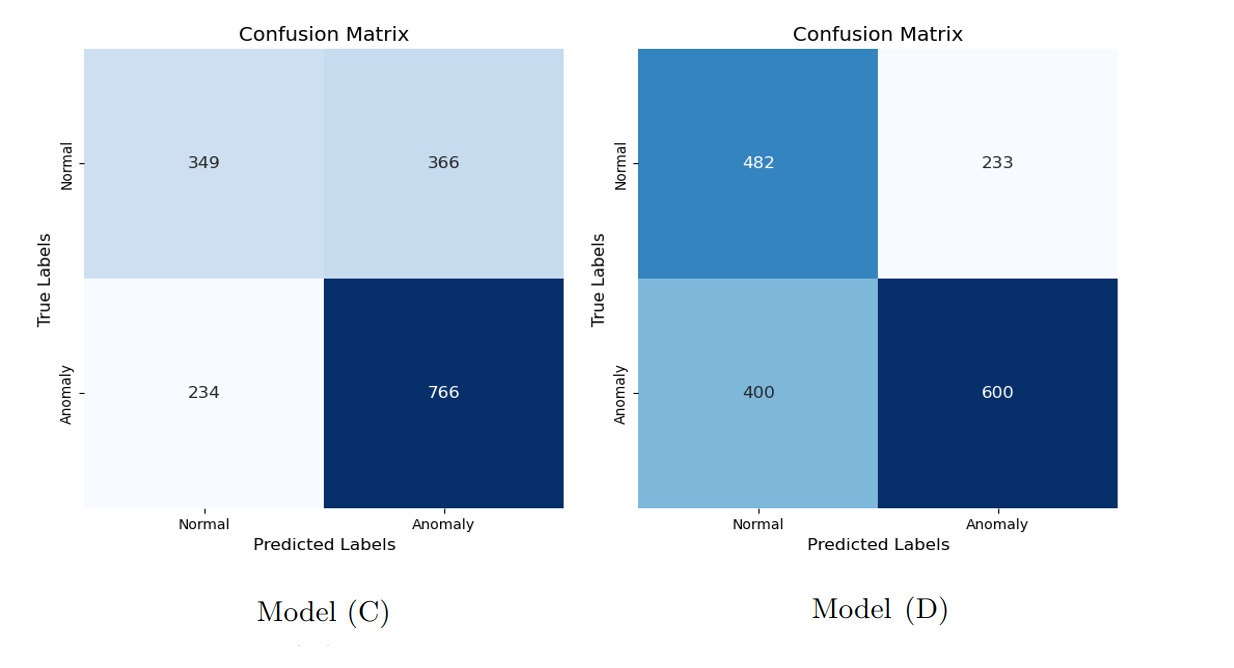} 
 \end{center}
\caption{Confusion matrix comparison for Model (C) and (D) for LOF}
\label{fig_f}
\end{figure}

In \ref{tab_h2} and Figure \ref{fig_f}, it is seen that Model (C) outperformed in recall (76.6\% vs 60\%) and F1-score (71.9\% vs 65.5\%), indicating its superiority in
detecting real anomalies. In contrast, while Model (D) has a higher precision of 72\%, it struggles to identify all real anomalies, as
revealed by its low recall of 60\%.

While drawing definitive conclusions on the influence of the number of neighbours from just two models is
challenging, there is a trend: models with better accuracies consistently have a higher number of neighbours in LOF. Similarly with isolation forest, the contamination parameter significantly drops LOF's performance. It typically
reflects the proportion of outliers in the dataset and was highly influential for both isolation forest and LOF models.

\subsubsection{One-Class SVM}

\begin{table}[!ht]
\caption{Parameter information and results for two models of one-class SVM}
    \centering
    \begin{tabular}{p{0.1\linewidth}  p{0.1\linewidth} p{0.1\linewidth} p{0.1\linewidth} p{0.1\linewidth} p{0.1\linewidth} p{0.1\linewidth} p{0.1\linewidth}}     \toprule
        Model & Kernel & Nu value & Accuracy & Precision & Recall & F1-score & MCC \\ \midrule
        Model (E) & RBF & 0.6 & 0.554 & 0.608 & 0.662 & 0.634 & 0.074 \\ %\hline
        Model (F) & RBF & 0.3 & 0.482 & 0.604 & 0.322 & 0.42 & 0.032 \\ \bottomrule
    \end{tabular}\label{tab_i}
\end{table}   

\begin{figure}[!ht]
\noindent
\begin{center}
 \includegraphics[width=1.0\textwidth]{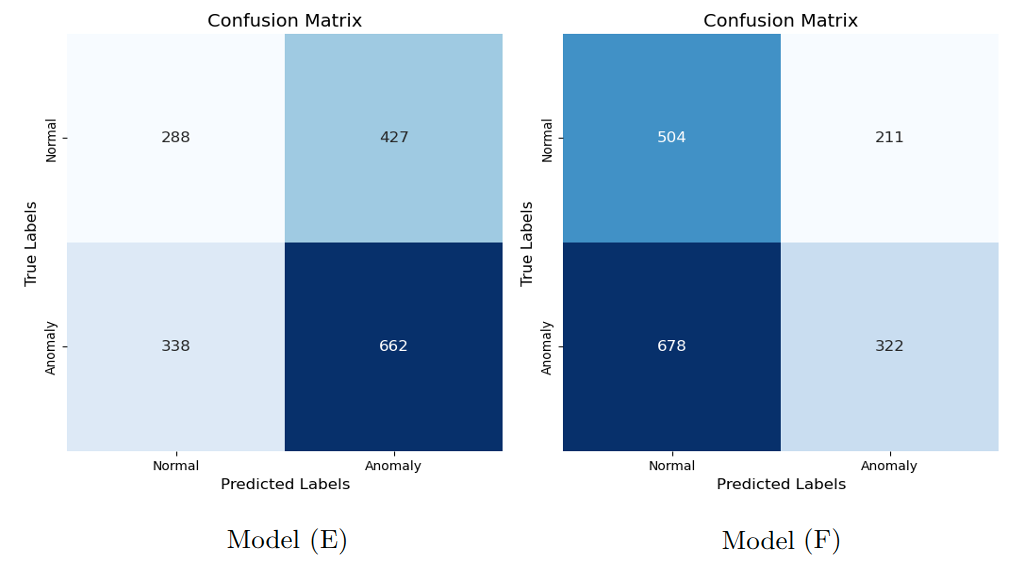} 
 \end{center}
\caption{Confusion matrix comparison for Model (E) and (F) for one-class SVM.}
\label{fig_f2}
\end{figure}

Both models, (E) and (F), employ the RDF kernel, allowing us to compare the performance metrics due to the variations in the Nu parameter Table \ref{tab_i}.  The difference in the Nu values significantly impacts the models' performances. Model (E), with a Nu value of 0.6, exhibits better overall performance across all metrics than Model (F), which has a Nu value of 0.3. Specifically, Model (F)'s performance lags, especially in terms of recall (32.2\%), which is considerably lower
than Model (E)'s recall (66.2\%) as in Figure \ref{fig_f2}.

\subsubsection{Visualization of the Results }
\begin{figure}[!ht]
\noindent
\begin{center}
 \includegraphics[width=0.88\textwidth]{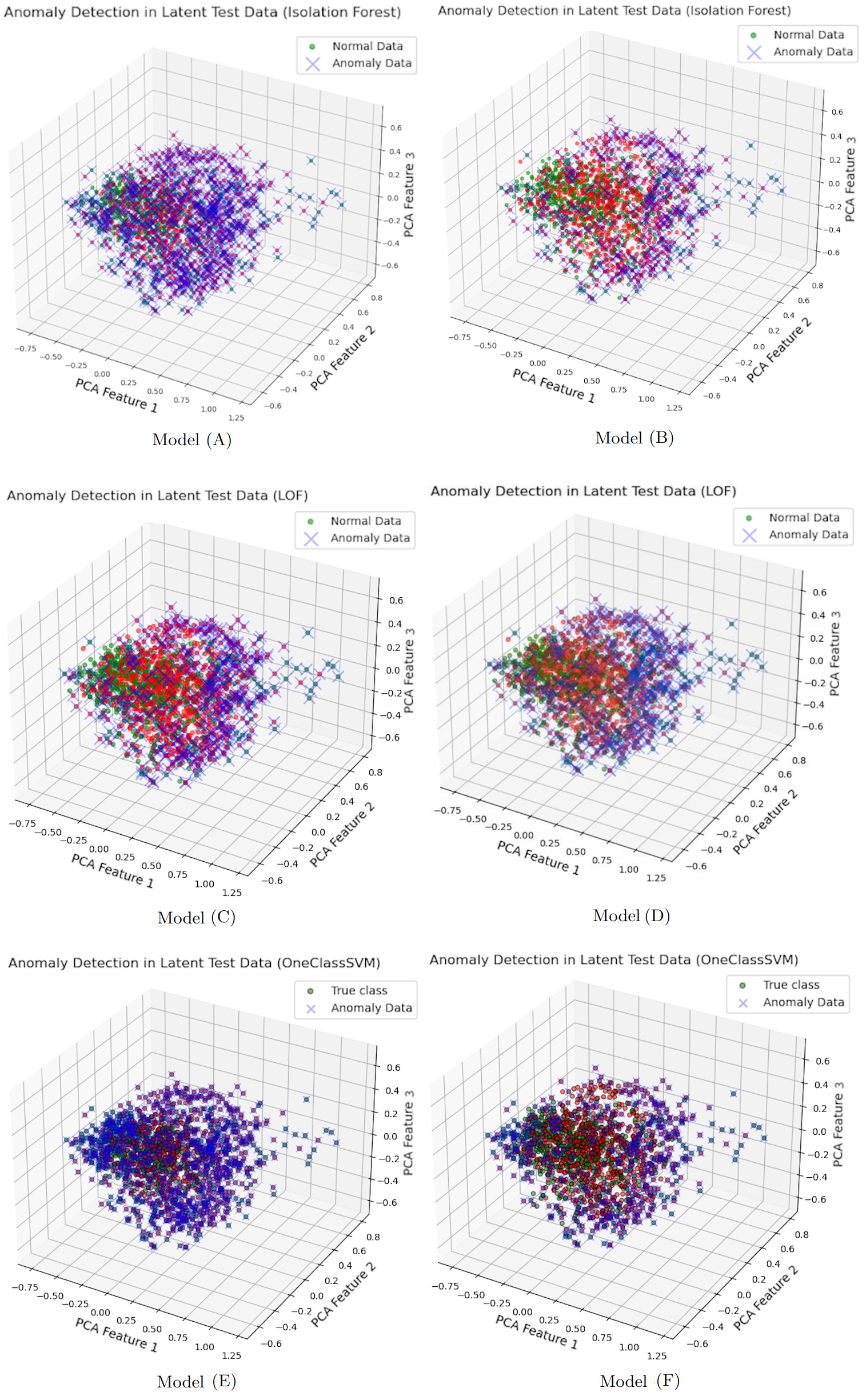} 
 \end{center}
\caption{Anomaly detection results of isolation forest on 3D maps (PCA is used for only visualization
purposes) for Model (A) and (B) (top row), (C) and (D) (middle row), and (E) and (F) (bottom row).}
 \label{fig_f3}
\end{figure}

Figure \ref{fig_f3} represents the anomaly detection results of all three techniques on 3D maps with PCA features. PCA features are created here just for visualisation purposes. The actual clustering, fitting,
and prediction processes utilise the original latent data generated through an autoencoder, detailed in Section
\ref{anom_clust}. Latent test data, which is 8-dimensional, is reduced into 3-dimensional data using PCA so that it is
possible to visualise the whole data in a 3D dimension plot instead of randomly choosing three variables.

Upon reviewing Figure \ref{fig_f3}, it is clear that anomaly and normal data points do not lie in the different spaces, which makes the clustering approach challenging. Furthermore, the significant reliance of these models on specific parameters like contamination and Nu is evident, as indicated in previous sections. While these approaches showed initial promise, they might not be the most suitable choice for imbalanced datasets like this one.

\subsection{Anomaly Detection Based on Reconstruction Error}

\subsubsection{LSTM autoencoder}

\begin{figure}[!ht]
\noindent
\begin{center}
 \includegraphics[width=1.0\textwidth]{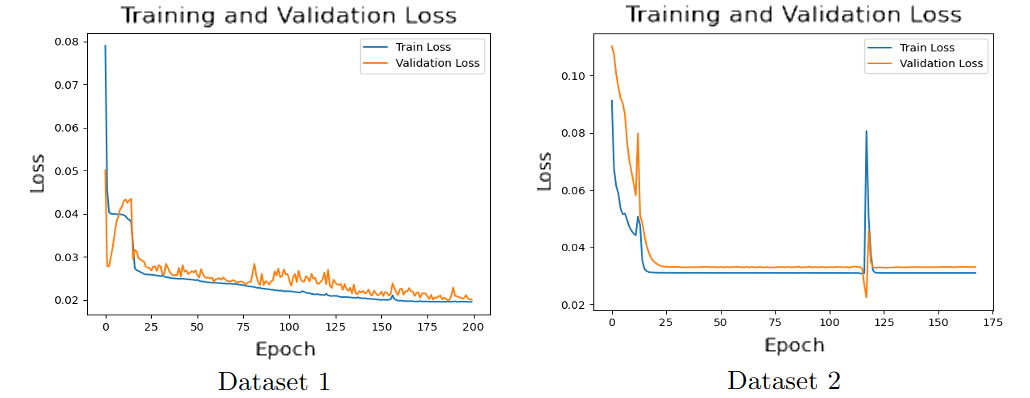} 
 \end{center}
\caption{LSTM Autoencoder training and validation loss plots. Dataset 1 on the left, Dataset 2 on the right.}
\label{fig_l}
\end{figure}

The model definition used for the training was presented in Table \ref{tab_gg2}. Both datasets training underwent training for 200 epochs,
incorporating an early stopping mechanism to mitigate overfitting. Figure \ref{fig_l} displays the training and validation
loss plots of the LSTM autoencoder for both Dataset 1 and Dataset 2. 

For Dataset 1, although the initial epochs show fluctuations, the model seems to stabilize and converge without any evident signs of overfitting or underfitting. Both the training and validation losses exhibit a consistent declining trend across the 200 epochs in Figure \ref{fig_l}. Early stopping is not triggered, suggesting that extending the training might further
reduce validation loss. For Dataset 2, the loss plots appear less consistent. Around the 125th epoch, there is a sudden spike in loss followed
by a rapid decline. Even as training continues to the 175th epoch, the model does not attain much additional learning. Table \ref{tab_j} presents the obtained performance metrics of models on the test data for Dataset 1 and Dataset 2.

\begin{table}[!ht]
\caption{Two LSTM autoencoder models based on Dataset 1 and Dataset 2, on test data.}
    \centering
    \begin{tabular}{p{0.2\linewidth}  p{0.12\linewidth}  p{0.12\linewidth}  p{0.12\linewidth}  p{0.12\linewidth}  p{0.12\linewidth}}     \toprule
         Model & Accuracy & Precision & Recall & F1-score & MCC \\ \midrule
        LSTM Autoencoder (Dataset 1) & 0.563 & 0.581 & 0.897 & 0.705 & -0.013 \\ %\hline
        LSTM Autoencoder (Dataset 2) & 0.457 & 0.526 & 0.692 & 0.598 &  -0.20 \\ \bottomrule
    \end{tabular}\label{tab_j}
\end{table}

\begin{figure}[!ht]
\noindent
\begin{center}
 \includegraphics[width=1.0\textwidth]{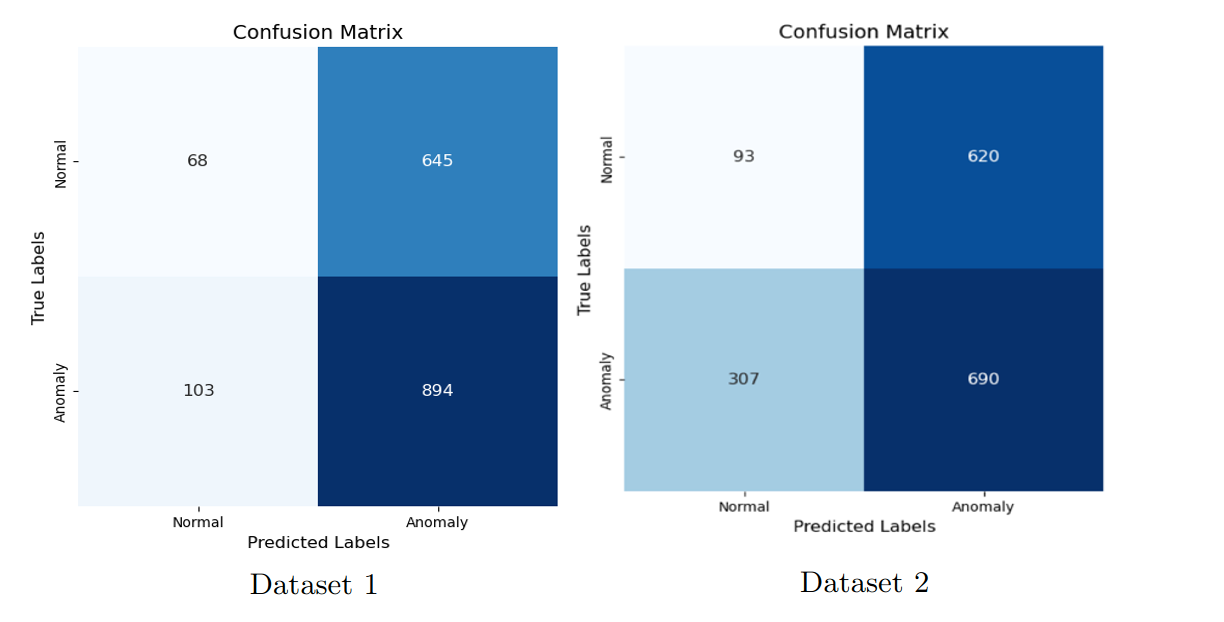} 
 \end{center}
\caption{Confusion matrices of LSTM autoencoder for Dataset 1 and Dataset 2.}
\label{fig_m}
\end{figure}

\begin{figure}[!ht]
\noindent
\begin{center}
 \includegraphics[width=1.0\textwidth]{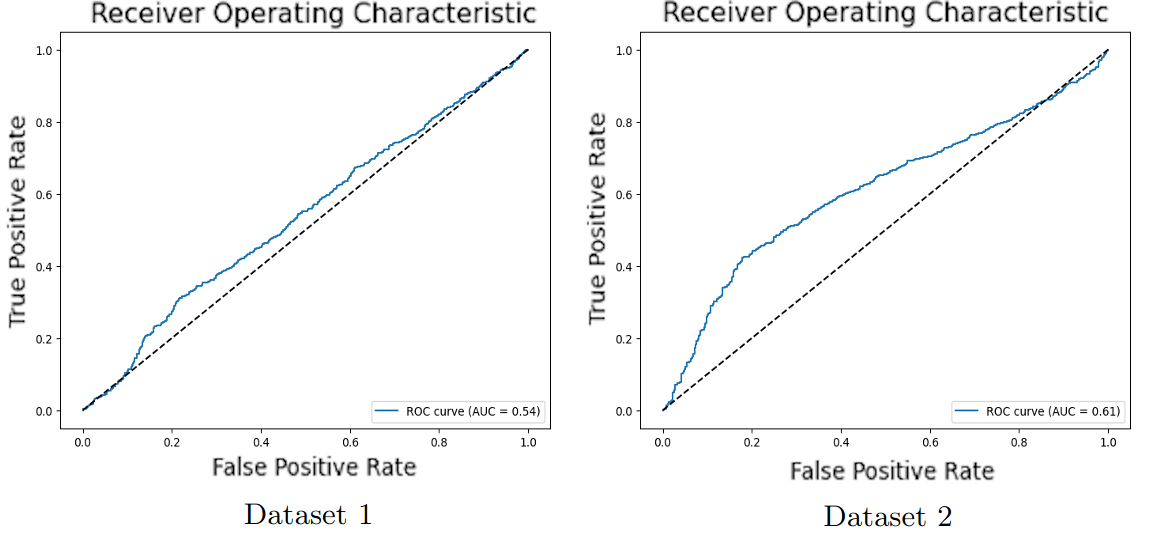} 
 \end{center}
\caption{ROC curve plots of LSTM autoencoder Dataset 1 and Dataset 2.}
\label{fig_n}
\end{figure}

From Table \ref{tab_j} and Figure \ref{fig_m}, it is observed that the model trained on Dataset 1 demonstrates better performance in
Accuracy, Precision, Recall, and F1-score compared to the model trained on Dataset 2, which retains only 17 variables. However, an important point for both models is that MCC is very low. While the first model's MCC is
closer to zero, suggesting a near-random prediction, the second model's MCC is significantly
lower, indicating a slightly worse prediction than random guessing. Analysing the ROC curve plots in Figure \ref{fig_n}, the second model exhibits a better AUC value of 0.61 than the first
model with 0.51. However, one could say that this type of ROC curve plot is not desirable.

In conclusion, while the first model, trained on Dataset 1, outperforms traditional performance metrics, the second
model's ROC curve seems better. Nevertheless, the overall quality of both models is not very
promising, given that an AUC close to 0.5 implies nearly random prediction. However, one can note that this reflects the inherent difficulty in predicting wildfires with limited historical data.

\subsubsection{FC Autoencoder}

The models A and B trained in this scenario were presented in Table \ref{tab_ii}. Each model's training process spanned 400 epochs, with an early stopping
(patience criteria as 20) integrated to avoid potential overfitting.

\begin{figure}[!ht]
\noindent
\begin{center}
 \includegraphics[width=1.0\textwidth]{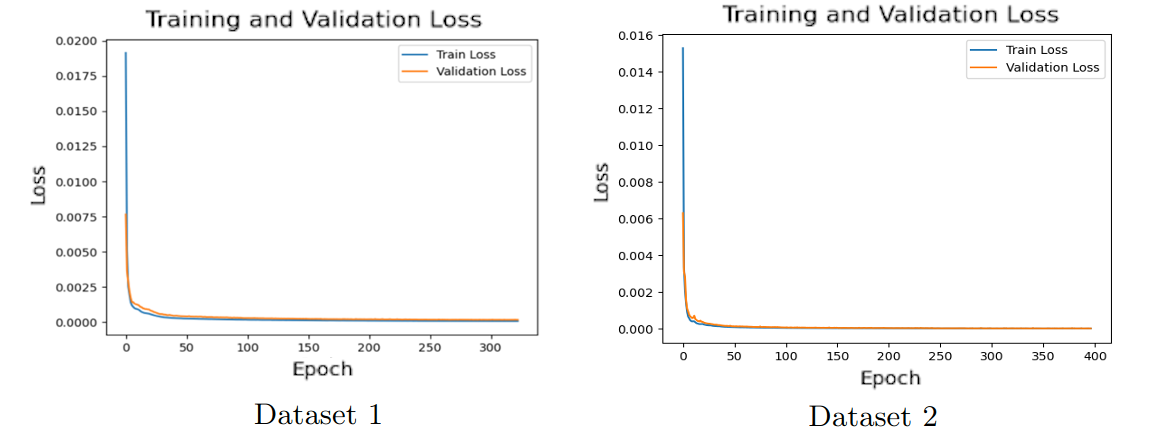} 
 \end{center}
\caption{FC Autoencoder (Model A) training and validation loss plots. Dataset 1 on the left, Dataset 2 on
the right.}
\label{fig_o}
\end{figure}

Based on Figure \ref{fig_o}, the FC Autoencoder Model A demonstrates steady convergence in its training and validation
loss plots for both Dataset 1 and Dataset 2. This smooth progression suggests effective learning without erratic
fluctuations, which is favourable. However, the early stopping could have been triggered earlier. This might indicate that the early stopping
criteria, the patience parameter, which is 20, in this case, were set too early, which allowed the model to train
longer than necessary without significant performance gains.

\begin{figure}[!ht]
\noindent
\begin{center}
 \includegraphics[width=1.0\textwidth]{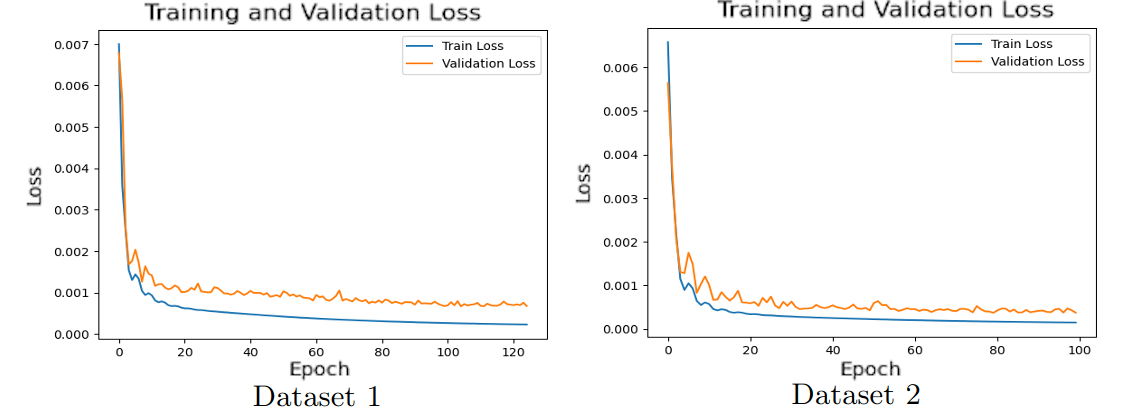} 
 \end{center}
\caption{FC Autoencoder (Model B) training and validation loss plots. Dataset 1 on the left, Dataset 2 on
the right.}
\label{fig_p}
\end{figure}

In Figure \ref{fig_p}, the FC Autoencoder Model B shows a less consistent convergence than Model A across both datasets.
Notable fluctuations in loss are observed, with early stopping triggered around the 120th epoch for Dataset 1 and the
100th epoch for Dataset 2. Additionally, a significant gap between training and validation losses in Dataset 1 suggests potential
overfitting, contrasting with the performance observed in Model B.

\begin{figure}[!ht]
\noindent
\begin{center}
 \includegraphics[width=1.0\textwidth]{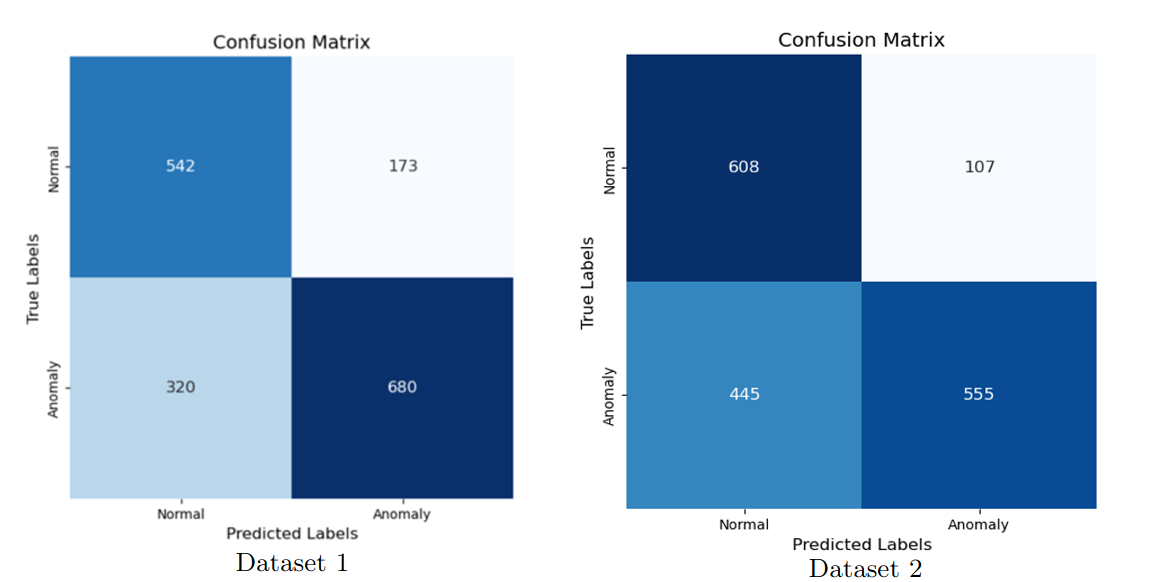} 
 \end{center}
\caption{Confusion matrices of FC Autoencoder (Model A) on the test data. Dataset 1 on the left, Dataset 2
on the right.}
\label{fig_q}
\end{figure}

\begin{figure}[!ht]
\noindent
\begin{center}
 \includegraphics[width=1.0\textwidth]{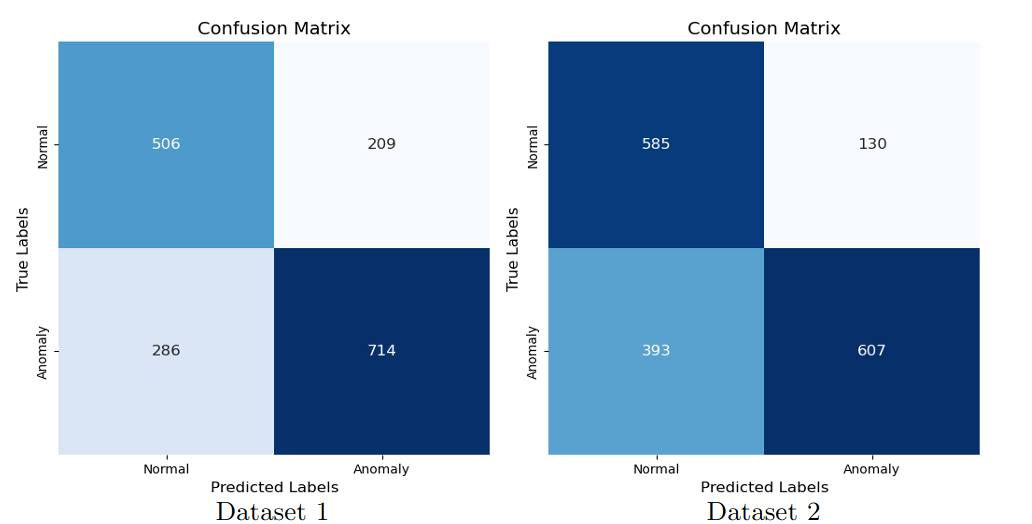} 
 \end{center}
\caption{Confusion matrices of FC Autoencoder (Model B) on the test data. Dataset 1 on the left, Dataset 2
on the right.}
\label{fig_r}
\end{figure}

Table \ref{tab_k} represents the obtained performance metrics of Models A and B on the test data of Dataset 1 and Dataset 2. Upon analysing the performance metrics from Table \ref{tab_k} and confusion matrices in Figures \ref{fig_q} and \ref{fig_r}, several insights emerge
regarding the efficacy of Models A and B across Datasets 1 and 2. For Dataset 1, both models show relatively similar performance metrics. However, a key distinction arises in the
recall, where Model B has a noticeable advantage with 71.4\% while Model A's recall is 68\%. On
the contrary, the MCC value of Model A is slightly higher than Model B, which may be due to its use of the Cyclical
Learning Rate scheduler, enabling more refined learning rate adjustments. For Dataset 2, although both models showcase very close precision, Model B compensates again with a better recall,
offering 60.7\% against Model A's 55.5\%. Model B also marginally outperforms Model A in accuracy
and MCC. 

\begin{table}[!ht]
\caption{Model A and B performance metrics on both test data of Dataset 1 and
Dataset 2.}
    \centering
    \begin{tabular}{p{0.15\linewidth}  p{0.15\linewidth} p{0.15\linewidth} p{0.15\linewidth} p{0.15\linewidth} p{0.15\linewidth}}    
    \toprule
        Model & Accuracy & Precision & Recall & F1-score & MCC \\ \midrule
         Model A (Dataset 1) & 0.712 & 0.797 & 0.68 & 0.733 & 0.432 \\ %\hline
        Model A (Dataset 2) & 0.678 & 0.838 & 0.555 & 0.667 & 0.41 \\ %\hline
         Model B (Dataset 1) & 0.711 & 0.773 & 0.714 & 0.742 & 0.417 \\ %\hline
        Model B (Dataset 2) & 0.695 & 0.823 & 0.607 & 0.698 & 0.423 \\ \bottomrule
    \end{tabular}\label{tab_k}
\end{table}

\begin{figure}[!ht]
\noindent
\begin{center}
 \includegraphics[width=1.0\textwidth]{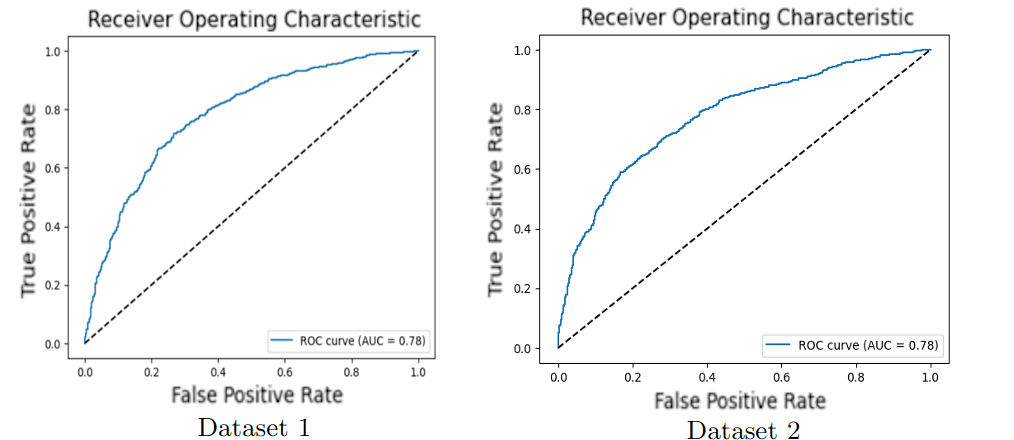} 
 \end{center}
\caption{ROC curve plots of FC Autoencoder (Model A). Dataset 1 on the left, Dataset 2 on the right.}
\label{fig_s}
\end{figure}

\begin{figure}[!ht]
\noindent
\begin{center}
 \includegraphics[width=1.0\textwidth]{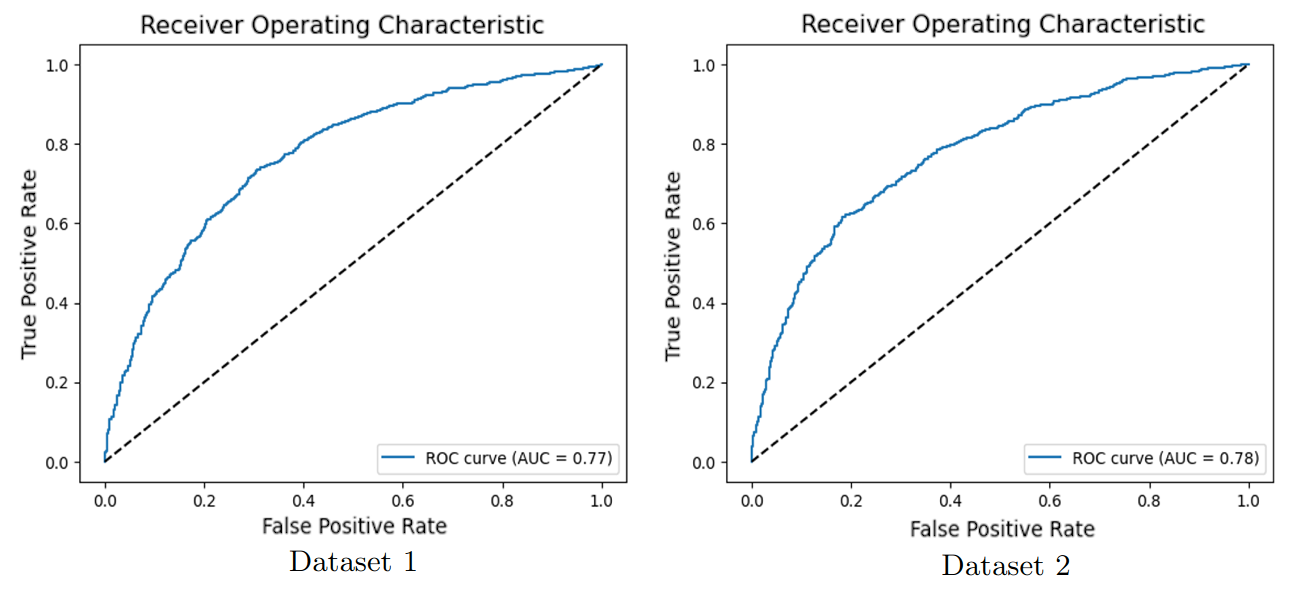} 
 \end{center}
\caption{ROC curve plots of FC Autoencoder (Model B). Dataset 1 on the left, Dataset 2 on the right.}
\label{fig_t}
\end{figure}

Figures \ref{fig_s} and \ref{fig_t} display the ROC curve plots for Models A and B across both Dataset 1 and Dataset 2 (using test data). All
these plots fall within the 77\% to 78\% range, making it difficult to draw meaningful comparisons between the models
and datasets. Nonetheless, the near-80\% performance indicates that the models consistently show quite promising
results across both datasets. 

Analysing the influence of datasets on model performance reveals some significant insights. Both models demonstrate
strong performance metrics even when applied to Dataset 2, which has been reduced to 17 variables. There is only a
slight decrease in metrics when the models transition from Dataset 1 to Dataset 2. An interesting observation is that
Model B's accuracy on Dataset 2 drops slightly from 71.1\% to 69.5\%, yet its AUC remains consistent at 78\%. This
consistency is particularly impressive given that the models' hyperparameters were specifically tuned for Dataset 1.
Naturally, one might expect better results on Dataset 1 because of this tuning. However, the minor decline in
performance on Dataset 2 suggests that this compact dataset effectively integrates the most influential variables.

In a direct comparison of results, the standout models still come from Dataset 1. Model A outperforms precision and MCC, whereas Model B is better in recall and F1-score. Given the importance of accurately detecting wildfires, a focus on
recall becomes crucial. Hence, in such a scenario, Model B, with its superior recall, may be the more advantageous
choice.

\section{Discussion}\label{discussion}
In this research, two different unsupervised learning approaches to detect wildfires in Australia with deep autoencoders have been implemented successfully.

The first approach was anomaly detection based on clustering through latent features using deep autoencoders. Here, firstly an FC autoencoder is trained and applied to reduce the data dimension from 28 to 8. Then, this latent
data is fed into three different clustering methods (isolation forest, LOF, and one-class SVM) to spot outliers. This approach presented several challenges. First, the performance metrics of all three methods were closely dependent on
specific parameters, especially the contamination and Nu values. For instance, the recall value for the isolation forest
model dropped sharply from 67.5\% to 33.7\% when contamination was decreased from
0.5 tom 0.2. Adjusting contamination and Nu values, while helpful, may only serve as a specific short-term solution. In
addition, considering that these techniques are implemented in an “unsupervised learning” way, it is important to
remember that a real-life scenario will not have ground truth labelled (wildfire or non-wildfire) data available for parameter
adjustment. Another challenge related to how the anomaly data spread out. Thanks to 3D map visualisations done using
PCA components, it was observed that anomaly data outliers did not cluster separately from the usual normal data points.
Consequently, due to these challenges and results, clustering techniques alone were not sufficient, robust, or
efficient in distinguishing anomalies in this wildfire dataset. This realisation triggered the exploration of
alternative strategies, guiding us to the second approach: anomaly detection using deep autoencoders,
primarily based on reconstruction error.

The second approach involved using deep autoencoders to detect anomalies by using the reconstruction error. The
approach involved training the deep autoencoder on normal data points and then using this trained model on the
reconstruction of test data. Anomalies have higher reconstruction errors while normal data points have low
errors. Two different deep autoencoder type were implemented: LSTM and FC.

In the LSTM autoencoder case, a final model architecture was determined after hyperparameter tuning and was applied to both
Dataset 1 and Dataset 2. The model trained on Dataset 1 showcased better performance metrics with an
accuracy of 56.3\%, whereas the model trained on Dataset 2 managed only 45.7\%. A consistent issue with both models was
their tendency to misclassify non-wildfire cases, often labelling them as wildfires. Ultimately, the performance of
both models is not sufficient, as their AUC values close to 0.5 suggest nearly random predictions. Our initial
assumption was that utilizing LSTM layers would be advantageous given the time series nature of the data, hoping to
capture temporal information. However, this assumption did not hold true in practice. Potential reasons include the
short sequence span of 10 days, which might not provide enough temporal context. Additionally, the daily aggregation of
the data could have lost valuable temporal patterns that LSTM layers might have captured. Lastly, one can note that low values of accuracy are expected given the challenging nature of the data and the lack of historical ground truth. Eventually, the goal of this study is not to filter out methods with poorer performance but to provide a comprehensive comparison. 

The final modelling phase employed an FC autoencoder. Two promising models obtained from hyperparameter tuning were
utilized on both Dataset 1 and Dataset 2. The first model leveraged the Adam optimizer with a learning rate scheduler,
while the second relied on RMSProp without any scheduler. Despite the compact 17-feature dimension of Dataset 2, the models
maintained strong metrics on this dataset as well. However, Dataset 1's models did perform better eventually. Most
impressively, the top model - using RMSProp without any scheduler - recorded an accuracy of 71.1\%, an F1-score of
74.2\%, and an MCC of 0.417. These figures are significantly higher than what is achieved with earlier methods,
highlighting the effectiveness of this approach and the use of FC layers in deep autoencoders.

\section{Conclusion}\label{concl}
In environmental science and disaster prevention, detection, and prediction of wildfires in time, especially in
vulnerable regions such as Australia, remains a crucial concern. This research successfully employed several
unsupervised learning techniques, in the context of anomaly detection with deep autoencoders, to predict wildfires in
Australia. This approach fills a gap in the field of wildfire prediction.

Our research explored two primary methods. The first method focused on anomaly detection using latent features
extracted from a deep autoencoder and demonstrated initial promising results. Utilizing the FC autoencoder enabled significant
dimensionality reduction, and when paired with clustering techniques like isolation forest, LOF, and one-class SVM,
yielded some encouraging results. However, while these methods showed potential, they were somewhat sensitive to
specific parameters, which could affect their reliability in real-life scenarios without labels. Furthermore, the
fact that anomaly data is not scattered as a different cluster, challenged these techniques even more.

The second method, using anomaly detection through reconstruction error, presented a clearer potential.
The LSTM autoencoder, despite its theoretical applicability
to time-series data, failed to yield the expected outcomes. Yet, the FC autoencoder showcased impressive results. The final
FC autoencoder outperformed its counterparts, achieving an accuracy of 71.1\%, an F1-score of 74.2\%, and an MCC of
0.417 on Dataset 1.

This work proposed wildfire prediction  by employing unsupervised learning approaches without the need for ground truth data of historical wildfires. Our findings demonstrate the feasibility of these methodologies in complex real-life scenarios linked to wildfire detection and filling a significant gap in the literature. The use of deep autoencoders for wildfire prediction shows the effectiveness of the method. Given the critical role of forests in the global ecosystem and the pressing need for efficient wildfire detection, this study aims to present a meaningful and practical solution in wildfire prediction, reinforcing the critical bridge between nature's preservation and technological innovation using AI.

Several recommendations could be given for future work for researchers working in this area. To begin, while the LSTM autoencoder used in this research was limited to 10-day sequences due to computational limitations, longer sequences, such as weekly or monthly intervals, may improve its performance. Second, given the dataset's temporal nature, Bi-LSTM autoencoders, which can capture bidirectional temporal information, might be explored as an alternative to the traditional LSTM. Furthermore, there is an opportunity for more extensive hyperparameter tuning to discover possible alternative optimum model configurations. Modification of the decision rule can provide a probability of wildfire instead of a detection outcome. Finally, given the encouraging improvements in the autoencoder area, experimenting with Variational Autoencoders (VAEs) might be a promising next step. These proposals can help to refine and
improve the predictive ability of the models in the wildfire prediction domain using unsupervised ML and extend the applicability of the wildfire prediction methods to a global scale.

\section*{Open Research Section}
The main data source is the dataset released by IBM for the Call for Code Spot Challenge for Wildfires competition \cite{IBM}. It is available at \url{https://github.com/Call-for-Code/Spot-Challenge-Wildfires}

The specific elements from the previous dataset used in this article have been processed from the following original sources:  ERA5-Land hourly data from 1950 to present dataset \cite{era5land} available at the Copernicus Climate Change Service (C3S) Climate Data Store (CDS)  \url{https://cds.climate.copernicus.eu/cdsapp#!/dataset/reanalysis-era5-land} ;  MOD13Q1 MODIS/Terra Vegetation Indices 16-Day L3 Global 250m SIN Grid V006 \cite{ndvi} available at NASA EOSDIS Land Processes Distributed Active Archive Center \url{https://lpdaac.usgs.gov/products/mod13q1v006/} ; and \cite{mcdl} available at NASA FIRMS  \url{https://www.earthdata.nasa.gov/learn/find-data/near-real-time/firms/mcd14dl-nrt} 
   
%\acknowledgments

\bibliography{biblio}

\end{document}